\documentclass[Afour,sageh,times]{sagej}

\usepackage{moreverb,url}
\usepackage{algorithm, setspace}
\usepackage{algpseudocode}
\usepackage{csquotes}
\usepackage{array,multirow}
\usepackage{tabularx}
\usepackage{adjustbox}
\usepackage{multicol}
\usepackage{subfigure}
\usepackage{xcolor}
\usepackage[colorlinks,bookmarksnumbered,allcolors=orange]{hyperref}
\usepackage{color, colortbl}

\definecolor{Gray}{gray}{0.9}

\newcommand\BibTeX{{\rmfamily B\kern-.05em \textsc{i\kern-.025em b}\kern-.08em
T\kern-.1667em\lower.7ex\hbox{E}\kern-.125emX}}

\newcommand{\dlnote}[1]%
 {\textcolor{blue}{\textbf{DL: #1}}}
 \newcommand{\ebnote}[1]%
 {\textcolor{red}{\textbf{EB: #1}}}
 \newcommand{\dsnote}[1]%
 {\textcolor{green}{\textbf{DS: #1}}}
 
\def\volumeyear{2020}

\usepackage{mathtools,fancyhdr,amsmath,amssymb,amsthm,enumitem}
\newcommand{\argmin}{\operatornamewithlimits{arg\,min}}
\newcommand{\argmax}{\operatornamewithlimits{arg\,max}}
\DeclarePairedDelimiter\abs{\lvert}{\rvert}
\DeclarePairedDelimiter\norm{\lVert}{\rVert}

\newcommand{\asymeq}{\stackrel{\boldsymbol{\cdot}}{=}}
\newcommand\mydots{\hbox to 1em{.\hss.\hss.}}

\newtheorem{theorem}{Theorem}

\newtheorem{example}{Example}

\begin{document}

\runninghead{B\i y\i k et al.}

\title{Learning Reward Functions from Diverse Sources of Human Feedback: Optimally Integrating Demonstrations and Preferences}

\author{Erdem B\i y\i k\affilnum{1}, Dylan P. Losey\affilnum{2}, Malayandi Palan\affilnum{2}, Nicholas C. Landolfi\affilnum{2}, Gleb Shevchuk\affilnum{2}, and Dorsa Sadigh\affilnum{1,2}}

\affiliation{\affilnum{1}{Department of Electrical Engineering, Stanford University}\\
\affilnum{2}{Department of Computer Science, Stanford University}}

\corrauth{Erdem B\i y\i k}

\email{ebiyik@stanford.edu}

\begin{abstract}
Reward functions are a common way to specify the objective of a robot. As designing reward functions can be extremely challenging, a more promising approach is to directly learn reward functions from human teachers. Importantly, data from human teachers can be collected either \emph{passively} or \emph{actively} in a variety of forms: passive data sources include \textit{demonstrations}, (e.g., kinesthetic guidance), whereas \textit{preferences} (e.g., comparative rankings) are actively elicited. Prior research has \textit{independently} applied reward learning to these different data sources. However, there exist many domains where multiple sources are complementary and expressive. Motivated by this general problem, we present a framework to \textit{integrate} multiple sources of information, which are either passively or actively collected from human users. In particular, we present an algorithm that first utilizes user demonstrations to initialize a belief about the reward function, and then actively probes the user with preference queries to zero-in on their true reward. This algorithm not only enables us combine multiple data sources, but it also informs the robot \textit{when} it should leverage each type of information. Further, our approach accounts for the human's \textit{ability} to provide data: yielding user-friendly preference queries which are also theoretically optimal. Our extensive simulated experiments and user studies on a Fetch mobile manipulator demonstrate the superiority and the usability of our integrated framework.

\end{abstract}


\maketitle

\section{Introduction}

When robots enter everyday human environments they need to understand how they should behave. Of course, humans know what the robot should be doing --- one promising direction is therefore for robots to \emph{learn} from human experts. Several recent deep learning works embrace this approach, and leverage human demonstrations to try and extrapolate the right robot behavior for interactive tasks.

In order to train a deep neural network, however, the robot needs access to a \emph{large amount} of interaction data. This makes applying such techniques challenging within domains where large amounts of human data are not readily available. Consider domains like robot learning and human-robot interaction in general: here the robot must collect diverse and informative examples of how a human wants the robot to act and respond \citep{choudhury2019utility}.

Imagine teaching an autonomous car to safely drive alongside human-driven cars. During training, you demonstrate how to merge in front of a few different vehicles. The autonomous car learns from these demonstrations, and tries to follow your examples as closely as possible. But when the car is deployed, it comes across an aggressive driver, who behaves differently than anything that the robot has seen before --- so that matching your demonstrations unintentionally causes the autonomous car to have an accident!

\begin{figure}[t]
\includegraphics[width=1.0\columnwidth]{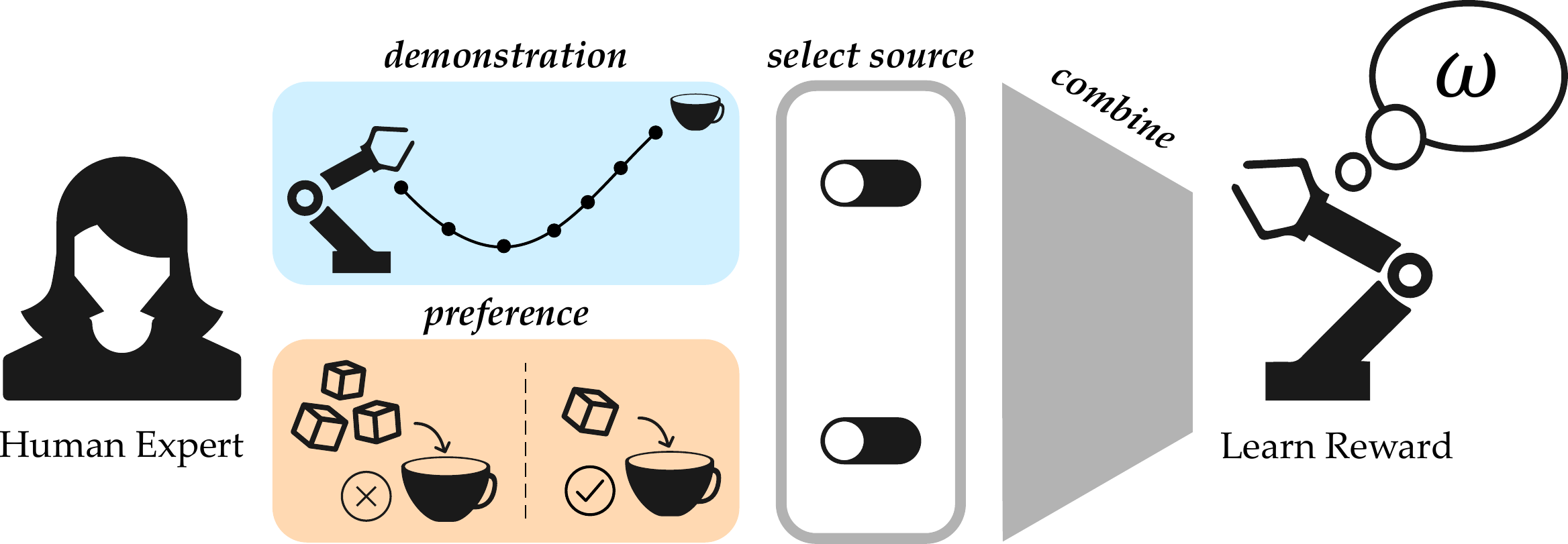}
\centering
\caption{When robots learn from human teachers multiple sources of information are available. We \emph{integrate} demonstrations and preferences, and determine \emph{when} to elicit \emph{what} type of data while accounting for the human-in-the-loop.}
\label{fig:overview}
\end{figure}

Here the autonomous car got it wrong because demonstrations alone failed to convey what you really wanted. Humans are unwilling --- and in many cases unable --- to provide demonstrations for every possible situation: but it is also difficult for robots to accurately generalize the knowledge learned from just a few expert demonstrations. Hence, the robot needs to leverage the time it has with the human teacher as intelligently as possible. Fortunately, there are many different \emph{sources} of human data: passive sources include demonstrations (e.g., kinesthetic teaching) and narrations, whereas preferences (e.g., rankings or queries) and verbal cues can be collected actively. All of these data sources provide information about the human's true reward function, but do so in different ways. Referring to Figure~\ref{fig:overview}, demonstrations provide rich and informative data about the style and timing of the robot's motions. On the other hand, preferences provide explicit and accurate information about a specific aspect of the task.

In this paper, we focus on leveraging these two sources, demonstrations and preferences, to learn what the user wants. Unlike prior work --- where the robot \emph{independently} learns from either source --- we assert that robots should intelligently \emph{integrate} these different sources when gathering data. Our insight is that:
\begin{center}
    \textit{Multiple data sources are complementary: demonstrations provide a high-level initialization of the human's overall reward functions, while preferences explore specific, fine-grained aspects of it.}
\end{center}
We present a unified framework for gathering human data that is collected either passively or actively. Importantly, our approach determines \emph{when} to utilize \emph{what} type of information source, so that the robot can learn efficiently. We draw from work on active learning and inverse reinforcement learning to synthesize human data sources while maximizing information gain.

We emphasize that the robot is gathering data from a human, and thus the robot needs to account for the human's \emph{ability} to provide that data. Returning to our driving example, imagine that an autonomous car wants to determine how quickly it should move while merging. There are a range of possible speeds from 20 to 30 mph, and the car has a uniform prior over this range. A na\"ive agent might ask a question to divide the potential speeds in half: would you rather merge at 24.9 or 25.1 mph? While another robot might be able to answer this question, to the human user these two choices seem indistinguishable --- rendering this question practically useless! Our approach ensures that the robot maintains a model of the human's \emph{ability to provide} data together with the \emph{potential value} of that data. This not only ensures that the robot learns as much as possible from each data source, but it also makes it \emph{easier} for the human to provide their data.

In this paper, we make the following contributions\footnote{Note that parts of this work have been published at Robotics: Science and Systems \citep{palan2019learning} and the Conference on Robot Learning \citep{biyik2019asking}.}:

\smallskip

\noindent \textbf{Determining When to Leverage Which Data Source.} We explore both passively collected demonstrations and actively collected preference queries. We prove that intelligent robots should collect passive data sources \textit{before} actively probing the human in order to maximize information gain. Moreover, when working with a human, each additional data point has an associated cost (e.g., the time required to provide that data). We therefore introduce an optimal stopping condition so that the robot \textit{stops} gathering data from the human when its expected utility outweighs its cost.

\smallskip

\noindent \textbf{Integrating Multiple Data Sources.} We propose a unified framework for reward learning, \textit{DemPref}, that leverages demonstrations and preferences to learn a personalized human reward function. We empirically show that combining both sources of human data leads to a better understanding of what the human wants than relying on either source alone. Under our proposed approach the human demonstrations initialize a high-level belief over what the right behavior is, and the preference queries iteratively fine-tune that belief to minimize robot uncertainty.

\smallskip

\noindent \textbf{Accounting for the Human.} When the robot tries to proactively gather data, it must account for the human's ability to provide accurate information. We show that today's state-of-the-art volume removal approach for generating preference queries does not produce easy or intuitive questions. We therefore propose an information theoretic alternative, which maximizes the utility of questions while readily minimizing the human's uncertainty over their answer. This approach naturally results in \textit{user-friendly} questions, and has the same computational complexity as the state-of-the-art volume removal method.

\smallskip

\noindent \textbf{Conducting Simulations and User Studies.} We test our proposed approach across multiple simulated environments, and perform two user studies on a 7-DoF Fetch robot arm. We experimentally compare our DemPref algorithm to (a) learning methods that only ever use a single source of data, (b) active learning techniques that ask questions without accounting for the human in-the-loop, and (c) algorithms that leverage multiple data sources but in different orders. We ultimately show that human end-users subjectively prefer our approach, and that DemPref objectively learns reward functions more efficiently than the alternatives.

\smallskip

Overall, this work demonstrates how robots can efficiently learn from humans by synthesizing multiple sources of data. We believe each of these data sources has a role to play in settings where access to data is limited, such as when learning from humans for interactive robotics tasks.

\section{Related Work}


Prior work has extensively studied learning reward functions using a single source of information, e.g., ordinal data \citep{chu2005gaussian}, or human corrections \citep{bajcsy2017learning,bajcsy2018learning,li2021learning}. Other works attempted to incorporate expert assessments of trajectories \citep{shah2020interactive}. More related to our work, we will focus on learning from demonstrations and learning from rankings. There has also been a few studies that investigate combining multiple sources of information. Below, we summarize these related works.

\smallskip

\noindent\textbf{Learning reward functions from demonstrations.} A large body of work is focused on learning reward functions using a single source of information: collected expert demonstrations. This approach is commonly referred to as inverse reinforcement learning (IRL), where the demonstrations are assumed to be provided by a human expert who is approximately optimizing the reward function~\citep{abbeel2004apprenticeship,abbeel2005exploration,ng2000algorithms,ramachandran2007bayesian,ziebart2008maximum,nikolaidis2015efficient}.

IRL has been successfully applied in a variety of domains. However, it is often too difficult to manually operate robots, especially manipulators with high degrees of freedom (DoF)~\citep{akgun2012keyframe,dragan2012formalizing,javdani2015shared,khurshid2015data}. Moreover, even when operating the high DoF of a robot is not an issue, people might have cognitive biases or habits that cause their demonstrations to not align with their actual reward functions. For example, \cite{kwon2020when} have shown that people tend to perform consistently risk-averse or risk-seeking actions in risky situations, depending on their potential losses or gains, even if those actions are suboptimal. As another example from the field of autonomous driving, \cite{basu2017you} suggest that people prefer their autonomous vehicles to be more timid compared to their own demonstrations. These problems show that, even though demonstrations carry an important amount of information about what the humans want, one should either try to learn from suboptimal demonstrations \citep{brown2020better,chen2020learning} or go beyond demonstrations to properly capture the underlying reward functions.

\smallskip

\noindent\textbf{Learning reward functions from rankings.}
Another helpful source of information that can be used to learn reward functions is rankings, i.e., when a human expert ranks a set of trajectories in the order of their preference~\citep{brown2019extrapolating,biyik2019green}. A special case of this, which we also adopt in our experiments, is when these rankings are pairwise \citep{akrour2012april,lepird2015bayesian,christiano2017deep,ibarz2018reward,brown2019deep}. In addition to the simulation environments, several works have leveraged pairwise comparison questions for various domains, including exoskeleton gait optimization \citep{tucker2020preference,li2021roial}, autonomous driving \citep{katz2021preference}, and trajectory optimization for robots in interactive settings \citep{cakmak2011human,biyik2020active}.

While having humans provide pairwise comparisons does not suffer from similar problems to collecting demonstrations, each comparison question is much less informative than a demonstration, since comparison queries can provide at most $1$ bit of information. Prior works have attempted to tackle this problem by actively generating the comparison questions \citep{sadigh2017active,biyik2018batch,basu2019active,katz2019learning,wilde2019bayesian}. While they were able to achieve significant gains in terms of the required number of comparisons, we hypothesize that one can attain even better data-efficiency by leveraging multiple sources of information, even when some sources might not completely align with the true reward functions, e.g., demonstrations as in the driving work by \cite{basu2017you}. In addition, these prior works did not account for the human in-the-loop and employed acquisition functions that produce very difficult questions for active question generation. In this work, we propose an alternative approach that generates easy comparison questions for the human while also maximizing the information gained from each question.
 
\smallskip

\noindent\textbf{Learning reward functions from both demonstrations and preferences.} \cite{ibarz2018reward} have explored combining demonstrations and preferences, where they take a model-free approach to learn a reward function in the Atari domain. Our motivation, physical autonomous systems, differs from theirs, leading us to a structurally different method. It is difficult and expensive to obtain data from humans controlling physical robots. Hence, model-free approaches are presently impractical. In contrast, we give special attention to data-efficiency. To this end, we (1) assume a simple reward structure that is standard in the IRL literature, and (2) employ active learning methods to generate comparison questions while simultaneously taking into account the ease of the questions. As the resulting training process is not especially time-intensive, we efficiently learn personalized reward functions.
\section{Problem Formulation}

\begin{figure}[t!]
\includegraphics[width=1.0\columnwidth]{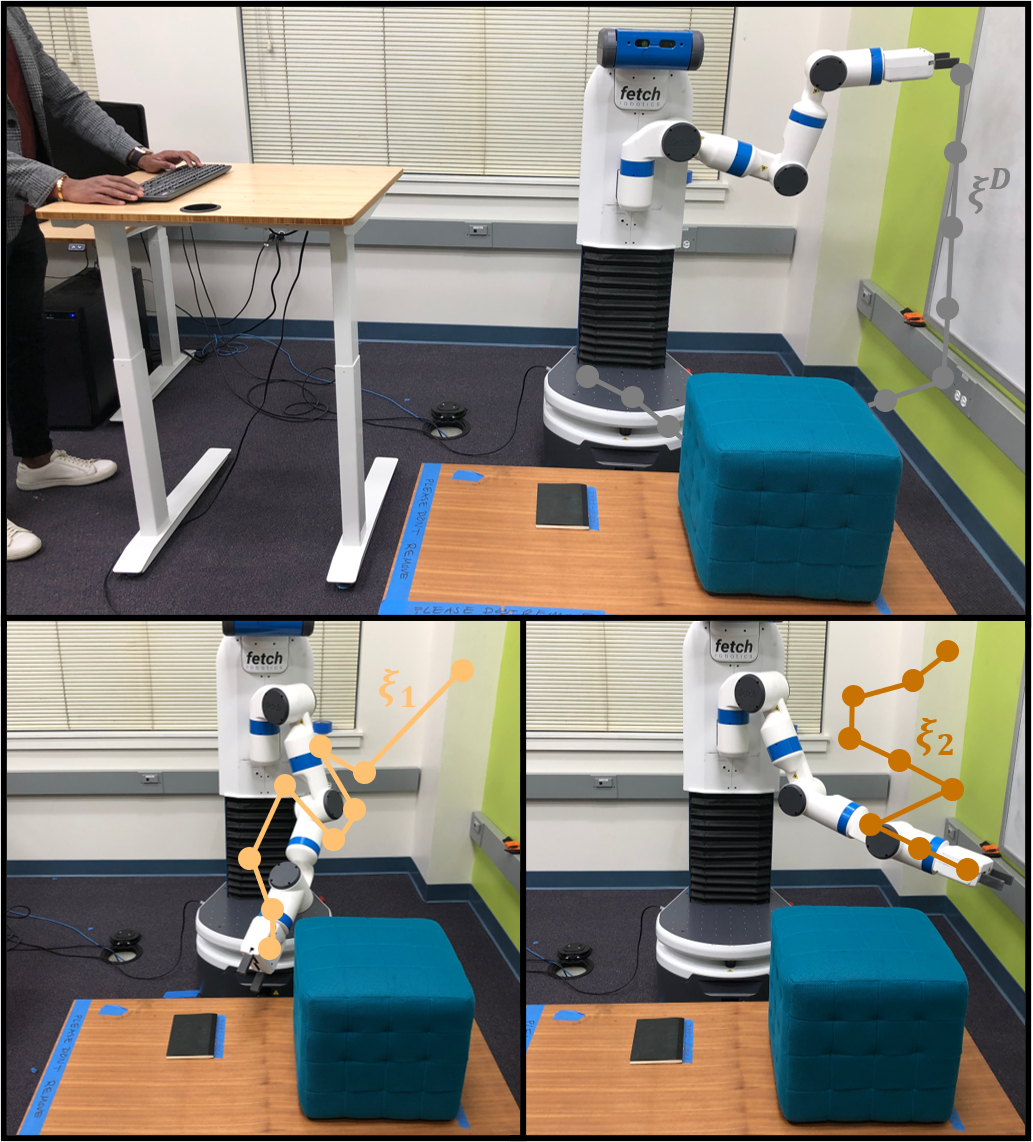}
\centering
\caption{Example of a demonstration (top) and preference query (bottom). During the demonstration the robot is \textit{passive}, and the human teleoperates the robot to produce trajectory $\xi^D$ from scratch. By contrast, the preference query is \textit{active}: the robot chooses two trajectories $\xi_1$ and $\xi_2$ to show to the human, and the human answers by selecting their preferred option.}
\label{fig:examples}
\end{figure}

\begin{figure*}[t]
\includegraphics[width=\textwidth]{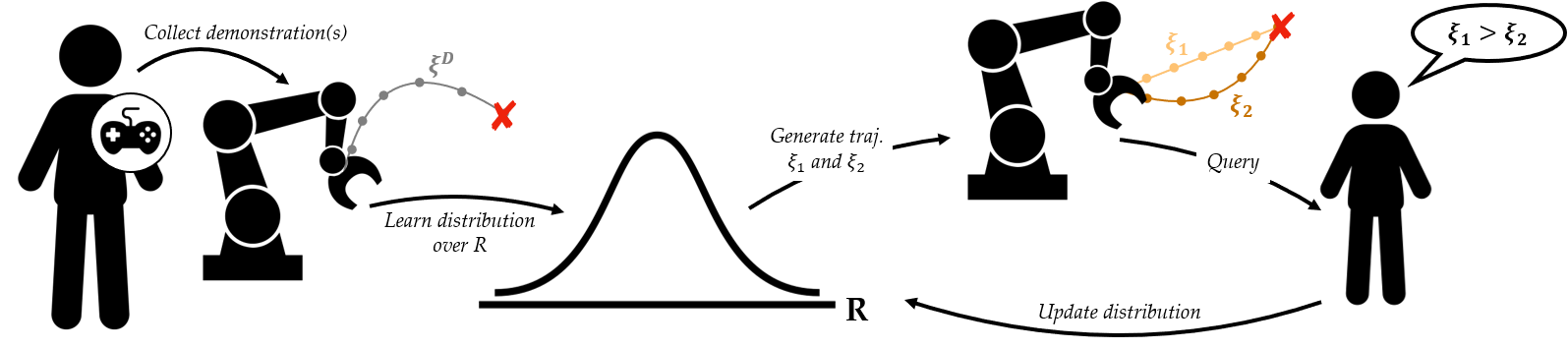}
\centering
\caption{Overview of our DemPref approach. The human starts by providing a set of \textit{high-level} demonstrations (left), which are used to initialize the robot's belief over the human's reward function. The robot then \textit{fine-tunes} this belief by asking questions (right): the robot actively generates a set of trajectories, and asks the human to indicate their favorite.}
\label{fig:dempref}
\end{figure*}

Building on prior work, we integrate multiple sources of data to learn the human's reward function. Here we formalize this problem setting, and introduce two forms of human feedback that we will focus on: demonstrations and preferences.

\smallskip

\noindent \textbf{MDP.} Let us consider a fully observable dynamical system describing the evolution of the robot, which should ideally behave according to the human's preferences. We formulate this system as a discrete-time Markov Decision Process (MDP) $\mathcal{M} = \langle \mathcal{S}, \mathcal{A}, f, r, T\rangle$. At time $t$, $s_t \in \mathcal{S}$ denotes the state of the system and $a_t \in \mathcal{A}$ denotes the robot's action. The robot transitions deterministically to a new state according to its dynamics: $s_{t+1} = f(s_t, a_t)$. At every timestep the robot receives reward $r(s)$, and the task ends after a total of $T$ timesteps.

\smallskip

\noindent \textbf{Trajectory.} A trajectory $\xi \in \Xi$ is a finite sequence of state-action pairs, i.e., $\xi = \big((s_t, a_t)\big)_{t=0}^T$ over time horizon $T$. Because the system is deterministic, the trajectory $\xi$ can be more succinctly represented by $\Lambda = (s_0, a_0, a_1, \ldots, a_T)$, the initial state and sequence of actions in the trajectory. We use $\xi$ and $\Lambda$ interchangeably when referring to a trajectory, depending on what is most appropriate for the context.

\smallskip

\noindent \textbf{Reward.} The reward function captures how the human wants the robot to behave. Similar to related works \citep{abbeel2004apprenticeship, ng2000algorithms, ziebart2008maximum}, we assume that the reward is a linear combination of features weighted by $\omega$, so that $r(s) = \omega \cdot \phi(s)$. The reward of a trajectory $\xi$ is based on the cumulative feature counts along that trajectory\footnote{More generally, the trajectory features $\Phi(\xi)$ can be defined as any function over the entire trajectory $\xi$.}:
\begin{equation} \label{eq:PF1}
    R(\xi) = \omega \cdot \sum_{s \in \xi} \phi(s) = \omega \cdot \Phi(\xi)
\end{equation}
Consistent with prior work \citep{bobu2018learning, bajcsy2017learning, ziebart2008maximum}, we will assume that the trajectory features $\Phi(\cdot)$ are given: accordingly, to understand what the human wants, the robot must simply learn the human's weights $\omega$. We normalize the weights so that $\|\omega\|_2 \leq 1$. 

\smallskip

\noindent \textbf{Demonstrations.} One way that the human can convey their reward weights to the robot is by providing demonstrations. Each human demonstration is a trajectory $\xi^D$, and we denote a data set of $L$ human demonstrations as $\mathcal{D} = \{\xi_1^D, \xi_2^D, \ldots, \xi_L^D\}$. In practice, these demonstrations could be provided by kinesthetic teaching, by teleoperating the robot, or in virtual reality (see Figure~\ref{fig:examples}, top).

\smallskip

\noindent \textbf{Preferences.} Another way the human can provide information is by giving feedback about the trajectories the robot shows. We define a preference query $Q = \{\xi_1, \xi_2, \ldots, \xi_K\}$ as a set of $K$ robot trajectories. The human answers this query by picking a trajectory $q \in Q$ that matches their personal preferences (i.e., maximizes their reward function). In practice, the robot could play $K$ different trajectories, and let the human indicate their favorite (see Figure~\ref{fig:examples}, bottom). 

\smallskip

\noindent \textbf{Problem.} Our overall goal is to accurately and efficiently learn the human's reward function from multiple sources of data. In this paper, we will only focus on demonstrations and preferences. Specifically, we will optimize when to query a user to provide demonstrations, and when to query a user to provide preferences. Our approach should learn the reward parameters $\omega$ with the smallest combination of demonstrations and preference queries. One key part of this process is accounting for the human-in-the-loop: we not only consider the informativeness of the queries for the robot, but we also ensure that the queries are intuitive and can easily be answered by the human user.
\section{DemPref: Learning Rewards from Demonstrations and Preferences}

In this section we overview our approach for integrating demonstrations and preferences to efficiently learn the human's reward function. Intuitively, demonstrations provide an informative, \textit{high-level} understanding of what behavior the human wants; however, these demonstrations are often noisy, and may fail to cover some aspects of the reward function. By contrast, preferences are \textit{fine-grained}: they isolate specific, ambiguous aspects of the human's reward, and reduce the robot's uncertainty over these regions. It therefore makes sense for the robot to start with high-level demonstrations before moving to fine-grained preferences. Indeed --- as we will show in Theorem~\ref{thm:order_of_dempref} --- starting with demonstrations and then shifting to preferences is the most efficient order for gathering data. Our DemPref algorithm leverages this insight to combine high-level demonstrations and low-level preference queries (see Figure~\ref{fig:dempref}).

\subsection{Initializing a Belief from Offline Demonstrations}

DemPref starts with a set of offline trajectory demonstrations $\mathcal{D}$. These demonstrations are collected \textit{passively}: the robot lets the human show their desired behavior, and does not interfere or probe the user. We leverage these passive human demonstrations to initialize an informative but imprecise prior over the true reward weights $\omega$.

\smallskip

\noindent \textbf{Belief.} Let the belief $b$ be a probability distribution over $\omega$. We initialize $b$ using the trajectory demonstrations, so that $b^0(\omega) = P(\omega  \mid  \mathcal{D})$. Applying Bayes' Theorem:
\begin{equation} \label{eq:DP1}
    \begin{split}
    b^0(\omega) & \propto P(\mathcal{D}  \mid  \omega)P(\omega) \\
    & \propto P(\xi^D_1, \xi^D_2, \ldots, \xi^D_L  \mid  \omega)P(\omega)
    \end{split}
\end{equation}
We assume that the trajectory demonstrations are conditionally independent, i.e., the human does not consider their previous demonstrations when providing a new demonstration. Hence, Equation (\ref{eq:DP1}) becomes:
\begin{equation} \label{eq:DP2}
    b^0(\omega) \propto P(\omega)\prod_{\xi^D \in \mathcal{D}} P(\xi^D  \mid  \omega)
\end{equation}
In order to evaluate Equation (\ref{eq:DP2}), we need a model of $P(\xi^D  \mid  \omega)$ --- in other words, how likely is the demonstrated trajectory $\xi^D$ given that the human's reward weights are $\omega$?

\smallskip

\noindent \textbf{Boltzmann Rational Model.} DemPref is not tied to any specific choice of the human model in Equation (\ref{eq:DP2}), but we do want to highlight the Boltzmann rational model that is commonly used in inverse reinforcement learning \citep{ziebart2008maximum, ramachandran2007bayesian}. Under this particular model, the probability of a human demonstration is related to the reward associated with that trajectory:
\begin{equation} \label{eq:DP3}
    \begin{split}
        P(\xi^D  \mid  \omega) & \propto \exp{\big(\beta^D R(\xi^D)\big)} \\
        & = \exp{\big(\beta^D\omega \cdot \Phi(\xi^D)\big)}
    \end{split}
\end{equation}
Here $\beta^D \geq 0$ is a temperature hyperparameter, commonly referred to as the \emph{rationality coefficient}, that expresses how noisy the human demonstrations are, and we substituted Equation (\ref{eq:PF1}) for $R$. Leveraging this human model, the initial belief over $\omega$ given the offline demonstrations becomes:
\begin{equation} \label{eq:DP4}
    b^0(\omega) \propto \exp{\Bigg(\beta^D\omega \cdot \sum_{\xi^D \in \mathcal{D}} \Phi(\xi^D)\Bigg)} P(\omega)
\end{equation}

\smallskip

\noindent \textbf{Summary.} Human demonstrations provide an informative but imprecise understanding of $\omega$. Because these demonstrations are collected passively, the robot does not have an opportunity to investigate aspects of $\omega$ that it is unsure about. We therefore leveraged these demonstrations to initialize $b^0$, which we treat as a high-level \textit{prior} over the human's reward. Next, we will introduce proactive questions to remove uncertainty and obtain a fine-grained posterior.

\subsection{Updating the Belief with Proactive Queries}

After initialization, DemPref iteratively performs two main tasks: \textit{actively} choosing the right preference query $Q$ to ask, and applying the human's answer to update $b$. In this section we focus on the second task: updating the robot's belief $b$. We will explore how robots should proactively choose the right question in the subsequent section.

\smallskip

\noindent \textbf{Posterior.} The robot asks a new question at each iteration $i\in \{0,\mydots\}$. Let $Q_i$ denote the $i$-th preference query, and let $q_i$ be the human's response to this query. Again applying Bayes' Theorem, the robot's posterior over $\omega$ becomes:
\begin{equation} \label{eq:DP5}
    b^{i+1}(\omega) \propto P(q_0,\mydots,q_i \mid Q_0, \mydots, Q_i, \omega) \cdot b^0(\omega),
\end{equation}
where $b^0$ is the prior initialized using human demonstrations. We assume that the human's responses $q$ are conditionally independent, i.e., only based on the current preference query and reward weights. Equation (\ref{eq:DP5}) then simplifies to:
\begin{equation} \label{eq:DP6}
    b^{i+1}(\omega) \propto \prod_{j=0}^i P(q_j \mid Q_j, \omega) \cdot b^0(\omega)
\end{equation}
We can equivalently write the robot's posterior over $\omega$ after asking $i+1$ questions as:
\begin{equation} \label{eq:DP7}
    b^{i+1}(\omega) \propto P(q_i \mid Q_i, \omega) \cdot b^i(\omega)
\end{equation}

\smallskip

\noindent \textbf{Human Model.} In Equation (\ref{eq:DP7}), $P(q \mid Q, \omega)$ denotes the probability that a human with reward weights $\omega$ will answer query $Q$ by selecting trajectory $q \in Q$. Put another way, this likelihood function is a probabilistic human model. Our DemPref approach is agnostic to the specific choice of $P(q \mid Q, \omega)$ --- we test different human models in our experiments. For now, we simply want to highlight that this human model defines the way users respond to queries.

\smallskip

\noindent \textbf{Choosing Queries.} We have covered how the robot can update its understanding of $\omega$ given the human's answers; but how does the robot choose the right questions in the first place? Unlike demonstrations --- where the robot is passive --- here the robot is \textit{active}, and purposely probes the human to get fine-grained information about specific parts of $\omega$ that are unclear. At the same time, the robot needs to remember that a human is answering these questions, and so the options need to be easy and intuitive for the human to respond to. Proactively choosing intuitive queries is the most challenging part of the DemPref approach. Accordingly, we will explore methods for actively generating queries $Q$ in the next section, before returning to finalize our DemPref algorithm.
\section{Asking Easy Questions}

Now that we have overviewed DemPref --- which integrates passive human demonstrations and proactive preference queries --- we will focus on how the robot chooses these queries. We take an \emph{active learning} approach: the robot selects queries to accurately fine-tune its estimate of $\omega$ in a data-efficient manner, minimizing the total number of questions the human must answer.

\smallskip

\noindent \textbf{Overview.} We present two methods for active preference-based reward learning: \emph{Volume Removal} and \emph{Information Gain}. Volume removal is a state-of-the-art approach where the robot solves a submodular optimization problem to choose which questions to ask.
However, this approach sometimes fails to generate \emph{informative} queries, and also does not consider the ease and intuitiveness of every query for the human-in-the-loop. This can lead to queries that are \emph{difficult} for the human to answer, e.g., two queries that are equally good (or bad) from the human's perspective.

We \emph{resolve} this issue with the second method, information gain: here the robot balances (a) how much information it will get from a correct answer against (b) the human’s ability to answer that question confidently. We end the section by describing a \emph{set of tools} that can be used to enhance either method, including an optimal condition for determining when the robot should stop asking questions.

\smallskip

\noindent \textbf{Greedy Robot.} The robot should ask questions that provide accurate, fine-grained information about $\omega$. Ideally, the robot will find the best \emph{adaptive sequence} of queries to clarify the human's reward. Unfortunately, reasoning about an adaptive sequence of queries is --- in general --- NP-hard \citep{ailon2012active}. We therefore proceed in a \emph{greedy} fashion: at every iteration $i$, the robot chooses $Q_i$ while thinking only about the next posterior $b^{i+1}$ in Equation (\ref{eq:DP7}).

\subsection{Choosing Queries with Volume Removal}

Maximizing volume removal is a state-of-the-art strategy for selecting queries. The method attempts to generate the most-informative queries by finding the $Q_i$ that maximizes the expected difference between the prior and \textit{unnormalized} posterior \citep{sadigh2017active, biyik2018batch, biyik2019green}. Formally, the method generates a query of $K\geq 2$ trajectories at iteration $i$ by solving:
\begin{equation*}
\argmax_{Q_i= \{\Lambda_1,\dots,\Lambda_K\}}\!\mathbb{E}_{q_i}\!\left[\int_{\norm{\omega}_2\leq 1}\!\left(b^i(\omega) - b^i(\omega)P(q_i \!\mid\! Q_i,\omega)\right)d\omega \right]
\end{equation*}
where the prior is on the left and the unnormalized posterior from Equation (\ref{eq:DP7}) is on the right.
This optimization problem can equivalently be written as:
\begin{equation} \label{eq:VR1}
Q^*_i = \argmax_{Q_i= \{\Lambda_1,\dots,\Lambda_K\}} \mathbb{E}_{q_i} \mathbb{E}_{b^i} \left[1 - P(q_i \!\mid\! Q_i,\omega)\right]
\end{equation}
The distribution $b^i$ can get very complex and thus --- to tractably compute the expectations in Equation (\ref{eq:VR1}) --- we are forced to leverage sampling. Letting $\Omega$ denote a set of $M$ samples drawn from the prior $b^i$, and $\asymeq$ denote asymptotic equality as the number of samples $M\to\infty$, the optimization problem in Equation (\ref{eq:VR1}) becomes:
\begin{equation} \label{eq:VR2}
Q^*_i \asymeq \argmin_{Q_i=\{\Lambda_1,\dots,\Lambda_K\}}\sum_{q_i\in Q_i}\left(\sum_{{\omega}\in\Omega}P(q_i\mid Q_i,{\omega})\right)^2
\end{equation}

\smallskip

\noindent \textbf{Intuition.} When solving Equation (\ref{eq:VR2}), the robot looks for queries $Q_i$ where each answer $q_i \in Q_i$ is equally likely given the current belief over $\omega$. These questions appear useful because the robot is maximally uncertain about which trajectory the human will prefer.

\smallskip

\noindent \textbf{When Does This Fail?} Although prior works have shown that volume removal can work in practice, we here identify two key shortcomings. First, we point out a failure case: the robot may solve for questions where the answers are equally likely but \textit{uninformative} about the human's reward. Second, the robot does not consider the human's ability to answer when choosing questions --- and this leads to \textit{challenging}, indistinguishable queries that are hard to answer!

\subsubsection{Uninformative Queries.} The optimization problem used to identify volume removal queries fails to capture our original goal of generating informative queries. Consider a trivial query where all options are identical: $Q_i = \{\xi_A, \xi_A, \mydots, \xi_A\}$. Regardless of which answer $q$ the human chooses, here the robot gets no information about the right reward function; put another way, $b^{i+1} = b^i$. Asking a trivial query is a waste of the human's time --- but we find that this uninformative question is actually a best-case solution to Equation (\ref{eq:VR1}).
\begin{theorem}
	The trivial query $Q=\{\xi_A, \xi_A, \mydots, \xi_A\}$ (for any $\xi_A\in\Xi$) is a global solution to Equation (\ref{eq:VR1}).
	\begin{proof}
	For a given $Q$ and $\omega$, $\sum_q P(q\mid Q,\omega) = 1$. Thus, we can upper bound the objective in Equation (\ref{eq:VR1}) as follows:
	\begin{align*}
		\mathbb{E}_{q_i} \mathbb{E}_{b^i} &[1 - P(q_i\mid Q_i,\omega)]\\
		&= 1 - \mathbb{E}_{b^i}\left[\sum_{q_i\in Q_i}P(q_i\mid Q_i,\omega)^2\right] \leq 1-1/K
	\end{align*}
	recalling that $K$ is the total number of options in $Q_i$. For the trivial query $Q = \{\xi_A, \xi_A, \mydots, \xi_A\}$, the objective in Equation~(\ref{eq:VR1}) has value $\mathbb{E}_{q_n} \mathbb{E}_{b^i} \left[1 - P(q_i\mid Q,\omega)\right]\ = 1 - 1/K$. This is equal to the upper bound on the objective, and thus the trivial, uninformative query of identical options is a global solution to Equation (\ref{eq:VR1}).\qed
	\end{proof}
\label{thm:volume_removal_failure}
\end{theorem}

\subsubsection{Challenging Queries.} Volume removal prioritizes questions where each answer is equally likely. Even when the options are not identical (as in a trivial query), the questions may still be very challenging for the user to answer. We explain this issue through a concrete example (also see Fig.~\ref{fig:corl19_domains}):

\begin{figure}[t]
\includegraphics[width=0.75\columnwidth]{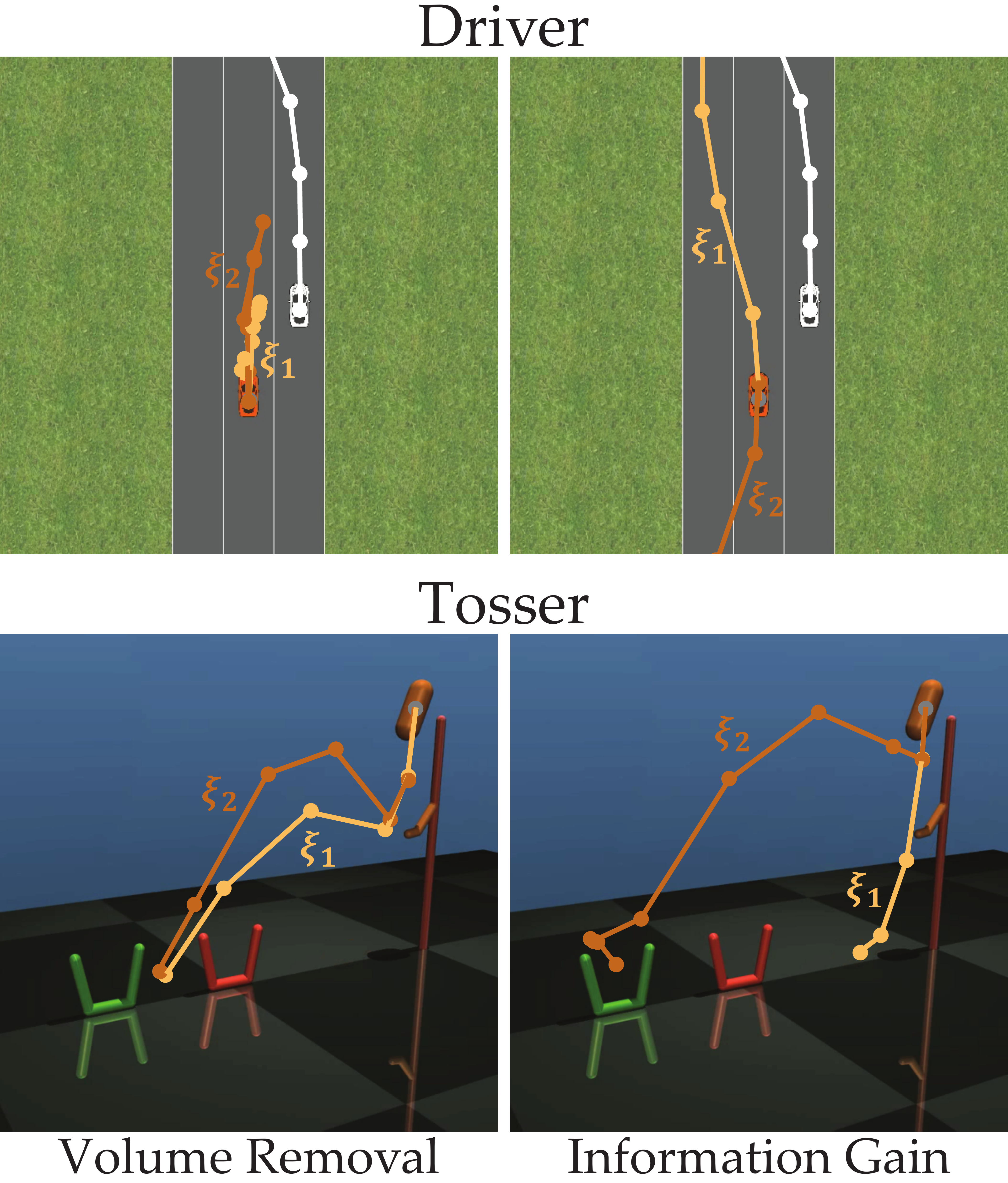}
\centering
\caption{Sample queries generated with the volume removal and information gain methods on Driver and Tosser tasks. Volume removal generates queries that are difficult, because the options are almost equally good or equally bad.}
\label{fig:corl19_domains}
\end{figure}

\begin{figure*}[t]
\includegraphics[width=\textwidth]{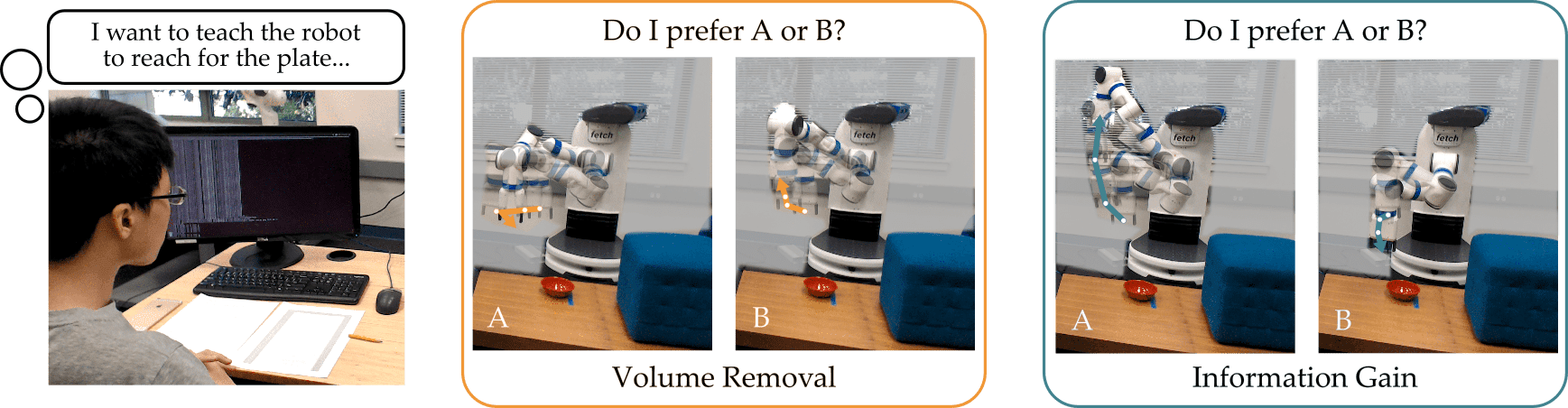}
\centering
\caption{Comparing preference queries that do not account for the human's ability to answer to queries generated using our information gain approach. Here the robot is attempting to learn the user’s reward function, and demonstrates two possible trajectories. The user should select the trajectory that better aligns with their own preferences. While the trajectories produced by the state-of-the-art volume removal method are almost indistinguishable, our information theoretic approach results in questions that are easy to answer, which eventually increase the robot’s overall learning efficiency.}
\label{fig:frontfig}
\end{figure*}

\begin{example}\label{ex:example_question}
	\normalfont Let the robot query the human while providing $K=2$ different answer options, $\xi_1$ and $\xi_2$.
	
	\textit{Question A.} Here the robot asks a question where both options are equally good choices. Consider query $Q_A$ such that $P(q = \xi_1 \mid Q_A, \omega) = P(q = \xi_2 \mid Q_A, \omega) \ \forall \omega\in\Omega$. Responding to $Q_A$ is difficult for the human, since both options $\xi_1$ and $\xi_2$ equally capture their reward function. 
	
	\textit{Question B.} Alternatively, this robot asks a question where only one option matches the human's true reward. Consider a query $Q_B$ such that:
	\begin{align*}
	    & P(q = \xi_1 \mid Q_B, \omega) \approx  1 \quad \forall \omega \in\Omega^{(1)} \\
	    & P(q = \xi_2 \mid Q_B, \omega) \approx  1 \quad \forall \omega \in\Omega^{(2)} \\
        & \Omega^{(1)} \cup \Omega^{(2)} = \Omega, \quad \abs{\Omega^{(1)}}=\abs{\Omega^{(2)}}
	\end{align*}
	If the human's weights $\omega$ lie in $\Omega^{(1)}$, the human will always answer with $\xi_1$, and --- conversely --- if the true $\omega$ lies in $\Omega^{(2)}$, the human will always select $\xi_2$. Intuitively, this query is easy for the human: regardless of what they want, one option stands out when answering the question.
\end{example}	

\noindent \textbf{Incorporating the Human.} Looking at Example~\ref{ex:example_question}, it seems clear that the robot should ask question $Q_B$. Not only does $Q_A$ fail to provide any information about the human's reward (because their response could be equally well explained by any $\omega$), but it is also hard for the human to answer (since both options seem equally viable). Unfortunately, when maximizing volume removal the robot thinks $Q_A$ is \textit{just as good} as $Q_B$: they are both global solutions to its optimization problem! Here volume removal gets it wrong because it fails to take the human into consideration. Asking questions based only on how uncertain the robot is about the human's answer can naturally lead to confusing, uninformative queries. Fig.~\ref{fig:corl19_domains} demonstrates some of these hard queries generated by the volume removal formulation.

\subsection{Choosing Queries with Information Gain}

To address the shortcomings of volume removal, we propose an information gain approach for choosing DemPref queries. We demonstrate that this approach ensures the robot not only asks questions which provide the most information about $\omega$, but these questions are also \textit{easy} for the human to answer. We emphasize that encouraging easy questions is not a heuristic addition: instead, the robot recognizes that picking queries which the human can accurately answer is necessary for overall performance. Accounting for the human-in-the-loop is therefore part of the optimal information gain solution.

\smallskip

\noindent \textbf{Information Gain.} At each iteration, we find the query $Q_i$ that maximizes the expected information gain about $\omega$. We do so by solving the following optimization problem:
\begin{align}\label{eq:IG1}
Q_i^* &= \argmax_{Q_i} I(\omega; q_i \mid Q_i, b^i) \nonumber\\
&= \argmax_{Q_i} H(\omega\mid Q_i, b^i) - \mathbb{E}_{q_i} H(\omega \mid q_i,Q_i, b^i),
\end{align}
where $I$ is the mutual information and $H$ is Shannon's information entropy \citep{cover2012elements}. Approximating the expectations via sampling, we re-write this optimization problem below (see the Appendix for the full derivation):
\begin{align}\label{eq:IG2}
Q^*_i \asymeq \argmax_{Q_i=\{\Lambda_1,\dots,\Lambda_K\}} &\frac{1}{M} \sum_{q_i\in Q_i}\sum_{{\omega}\in\Omega}\Bigg(P(q_i \mid Q_i,{\omega}) \nonumber\\
&\log_2{\!\left(\!\frac{M \cdot P(q_i\!\mid\! Q_i,{\omega})}{\sum_{\omega'\in\Omega}\!P(q_i \!\mid\!Q_i,\omega')}\!\right)\!\Bigg)}
\end{align}

\smallskip

\noindent\textbf{Intuition.} To see why accounting for the human is naturally part of the information gain solution, re-write Equation (\ref{eq:IG1}):
\begin{equation} \label{eq:IG3}
    Q_i^* = \argmax_{Q_i} ~ H(q_i \mid Q_i,b^i) - \mathbb{E}_{\omega} H(q_i \mid \omega, Q_i)
\end{equation}
Here the first term in Equation (\ref{eq:IG3}) is the \emph{robot's uncertainty} over the human's response: given a query $Q_i$ and the robot's understanding of $\omega$, how confidently can the robot predict the human's answer? The second entropy term captures the \emph{human's uncertainty} when answering: given a query and their true reward, how confidently will they choose option $q_i$? Optimizing for information gain with Equations (\ref{eq:IG1}) or (\ref{eq:IG2}) naturally considers both robot and human uncertainty, and favors questions where (a) the robot is unsure how the human will answer but (b) the human can answer easily. We contrast this to volume removal, where the robot purely focused on questions where the human's answer was unpredictable.

\smallskip

\noindent \textbf{Why Does This Work?} To highlight the advantages of this method, let us revisit the shortcomings of volume removal. Below we show how information gain successfully addresses the problems described in Theorem~\ref{thm:volume_removal_failure} and Example~\ref{ex:example_question}. Further, we emphasize that the computational complexity of computing objective (\ref{eq:IG2}) is equivalent --- in order --- to the volume removal objective from Equation (\ref{eq:VR2}). Thus, information gain avoids the previous failures while being at least as computationally tractable.

\subsubsection{Uninformative Queries.} Recall from Theorem~\ref{thm:volume_removal_failure} that any trivial query $Q = \{\xi_A,\ldots, \xi_A\}$ is a global solution for volume removal. In reality, we know that this query is a worst-case choice: no matter how the human answers, the robot will gain no insight into $\omega$. Information gain  ensures that the robot will not ask trivial queries: under Equation (\ref{eq:IG1}), $Q = \{\xi_A,\ldots, \xi_A\}$ is actually the \emph{global minimum}!

\subsubsection{Challenging Questions.} Revisiting Example~\ref{ex:example_question}, we remember that $Q_B$ was a much easier question for the human to answer, but volume removal values $Q_A$ as highly as $Q_B$. Under information gain, the \emph{robot} is equally uncertain about how the human will answer $Q_A$ and $Q_B$, and so the first term in Equation (\ref{eq:IG3}) is the same for both. But the information gain robot additionally recognizes that the \emph{human} is very uncertain when answering $Q_A$: here $Q_A$ attains the global maximum of the second term while $Q_B$ attains the global minimum! Thus, the overall value of $Q_B$ is higher and --- as desired --- the robot recognizes that $Q_B$ is a better question.

\smallskip

\noindent \textbf{Why Demonstrations First?} Now that we have a user-friendly strategy for generating queries, it remains to determine \emph{in what order} the robot should leverage the demonstrations and the preferences.

Recall that demonstrations provide coarse, high-level information, while preference queries hone-in on isolated aspects of the human's reward function. Intuitively, it seems like we should start with high-level demonstrations before probing low-level preferences: but is this really the right order of collecting data? What about the alternative --- a robot that instead waits to obtain demonstrations until after asking questions? 

When leveraging information gain to generate queries, we here prove that the robot will gain \emph{at least as much} information about the human's preferences as any other order of demonstrations and queries. Put another way, starting with demonstrations \emph{in the worst case} is just as good as any other order; and \emph{in the best case} we obtain more information.

\begin{theorem}
	Under the Boltzmann rational human model for demonstrations presented in Equation~\eqref{eq:DP3}, our DemPref approach (Algorithm~\ref{alg:DemPref}) --- where preference queries are actively generated after collecting demonstrations --- results in at least as much information about the human's preferences as would be obtained by reversing the order of queries and demonstrations.
	\begin{proof}
	Let $Q_i^*$ be the information gain query \emph{after collecting demonstrations}. From Equation (\ref{eq:IG1}), $Q_i^* = \argmax_{Q_i} I(\omega; q_i \mid Q_i, b^i)$. We let $q_i^*$ denote the human's response to query $Q_i^*$. Similarly, let $\tilde{Q}_i$ be the information gain query \emph{before collecting demonstrations}, so that $\tilde{Q}_i = \argmax_{Q_i} I(\omega; q_i \mid Q_i, (\tilde{Q}_j, \tilde{q}_j)_{j=0}^{i-1})$. Again, we let $\tilde{q}_i$ denote the human's response to query $\tilde{Q}_i$. We can now compare the overall information gain for each order of questions and demonstrations:
	\begin{align*}
	    I\Big(&\omega;\left(\mathcal{D},q^*_1,q^*_2,\mydots) \mid (Q^*_1,Q^*_2,\mydots)\right)\Big)\\
	    &= I(\omega;\mathcal{D}) + I\left(\omega;(q^*_1,q^*_2,\mydots) \mid (b^0, Q^*_1,Q^*_2,\mydots)\right)\\
	    &\geq I(\omega;\mathcal{D}) + I\left(\omega;(\tilde{q}_1,\tilde{q}_2,\mydots) \mid (b^0, \tilde{Q}_1,\tilde{Q}_2,\mydots)\right)\\
	    &= I\left(\omega;(\tilde{q}_1,\tilde{q}_2,\mydots, \mathcal{D}) \mid (\tilde{Q}_1,\tilde{Q}_2,\mydots)\right)\qquad\qquad\qed
	\end{align*}
	\end{proof}
\label{thm:order_of_dempref}
\end{theorem}

\noindent \textbf{Intuition.} We can explain Theorem~\ref{thm:order_of_dempref} through two main insights. First, the information gain from a passively collected demonstration is the same regardless of when that demonstration is provided. Second, proactively generating questions based on a prior leads to more incisive queries than choosing questions from scratch. In fact, Theorem~\ref{thm:order_of_dempref} can be generalized to show that active information resources should be utilized after passive resources.

\smallskip

\noindent \textbf{Bounded Regret.} At the start of this section we mentioned that --- instead of looking for the optimal sequence of future questions --- our DemPref robot will greedily choose the best query at the current iteration. Prior work has shown that this greedy approach is reasonable for volume removal, where it is guaranteed to have bounded sub-optimality in terms of the volume removed~\citep{sadigh2017active}. However, this volume is defined in terms of the unnormalized distribution, and so this result does not say much about the learning performance. Unfortunately, the information gain also does not provide theoretical guarantees, as it is only submodular, but not \emph{adaptive} submodular.

\subsection{Useful Tools \& Extensions}

We introduced how robots can generate proactive questions to maximize volume removal or information gain. Below we highlight some additional tools that designers can leverage to improve the computational performance and applicability of these methods. In particular, we draw the reader's attention to an optimal stopping condition, which tells the DemPref robot when to stop asking the human questions.

\smallskip

\noindent \textbf{Optimal Stopping.} We propose a novel extension --- specifically for information gain --- that tells the robot when to stop asking questions. Intuitively, the DemPref querying process should end when the robot's questions become more costly to the human than informative to the robot. 

Let each query $Q$ have an associated cost $c(Q) \in \mathbb{R}^+$. This function captures the \textit{cost} of a question: e.g., the amount of time it takes for the human to answer, the number of similar questions that the human has already seen, or even the interpretability of the question itself. We subtract this cost from our information gain objective in Equation (\ref{eq:IG1}), so that the robot maximizes information gain while biasing its search towards low-cost questions:
\begin{align} \label{eq:EX1}
\max_{Q_i=\{\Lambda_1,\mydots,\Lambda_K\}} I(\omega; q_i \mid Q_i, b^i) - c(Q_i)
\end{align}
Importantly, Theorem~\ref{thm:order_of_dempref} still holds with this extension to the model, since the costs depend only on the query. And now that we have introduced a cost into the query selection problem, the robot can reason about when its questions are becoming prohibitively expensive or redundant. We find that the best time to stop asking questions in expectation is when their cost exceeds their value:
\begin{theorem}\label{thm:optimal_stopping}
	A robot using information gain to perform active preference-based learning should stop asking questions if and only if the global solution to Equation \eqref{eq:EX1} is negative at the current iteration.
\end{theorem}
\noindent See the Appendix for our proof. We emphasize that this result is valid only for information gain, and adapting Theorem~\ref{thm:optimal_stopping} to volume removal is not trivial.

The decision to terminate our DemPref algorithm is now fairly straightforward. At each iteration $i$, we search for the question $Q_i$ that maximizes the trade-off between information gain and cost. If the value of Equation (\ref{eq:EX1}) is non-negative, the robot shows this query to the human and elicits their response; if not, the robot cannot find any sufficiently important questions to ask, and the process ends. This automatic stopping procedure makes the active learning process more user-friendly by ensuring that the user does not have to respond to unnecessary or redundant queries.

\smallskip

\noindent \textbf{Other Potential Extensions.} We have previously developed several tools to improve the computational efficiency of volume removal, or to extend volume removal to better accommodate human users. These tools include batch optimization \citep{biyik2018batch,biyik2019batch}, iterated correction \citep{palan2019learning}, and dynamically changing reward functions \citep{basu2019active}. Importantly, the listed tools are \textit{agnostic} to the details of volume removal: they simply require (a) the query generation algorithm to operate in a greedy manner while (b) maintaining a belief over $\omega$. Our proposed information gain approach for generating easy queries satisfies both of these requirements. Hence, these prior extensions to volume removal can also be straightforwardly applied to information gain.
\section{Algorithm}

\begin{algorithm}[t]
  \caption{DemPref with a Human-in-the-Loop}
  \label{alg:DemPref}
  \begin{algorithmic}[1]
    \State Collect human demonstrations: $\mathcal{D} = \{\xi^D_1, \xi^D_2, \ldots, \xi^D_L \}$
    \State Initialize belief over the human's reward weights $\omega$:
    \begin{equation*}
        b^0(\omega) \propto \exp{\Bigg(\beta^D\omega \cdot \sum_{\xi^D \in \mathcal{D}} \Phi(\xi^D)\Bigg)} P(\omega)
    \end{equation*}
    \For{$i \gets 0, 1, \ldots$}
        \State Choose proactive question $Q_i$:
        \begin{equation*}
            Q_i \gets \argmax_{Q}~I(\omega; q \mid Q, b^i) - c(Q)
        \end{equation*}
        \If{$I(\omega; q \mid Q_i, b^i) - c(Q_i) < 0$}
            \State \Return $b^i$
        \EndIf
        \State Elicit human's answer $q_i$ to query $Q_i$
        \State Update belief over $\omega$ given query and response:
        \begin{equation*}
            b^{i+1}(\omega) \propto P(q_i \mid Q_i, \omega) b^i(\omega)
        \end{equation*}
    \EndFor
  \end{algorithmic}
\end{algorithm}

We present the complete DemPref pseudocode in Algorithm~\ref{alg:DemPref}. This algorithm involves two main steps: first, the robot uses the human's offline trajectory demonstrations $\mathcal{D}$ to initialize a high-level understanding of the human's preferred reward. Next, the robot actively generates user-friendly questions $Q$ to fine-tune its belief $b$ over $\omega$. These questions can be selected using volume removal or information gain objectives (we highlight the information gain approach in Algorithm~\ref{alg:DemPref}). As the robot asks questions and obtains a precise understanding of what the human wants, the informative value of new queries decreases: eventually, asking new questions becomes suboptimal, and the DemPref algorithm terminates.

\smallskip

\noindent \textbf{Advantages.} We conclude our presentation of DemPref by summarizing its two main contributions:
\begin{enumerate}
    \item The robot learns the human's reward by synthesizing two types of information: high-level demonstrations and fine-grained preference queries.
    \item The robot generates questions while accounting for the human's ability to respond, naturally leading to user-friendly and informative queries.
\end{enumerate}
\section{Experiments} \label{sec:experiments}

We conduct five sets of experiments to assess the performance of DemPref under various metrics\footnote{Unless otherwise noted, we adopt $\beta^D=0.02$, constant $c(Q)$ for $\forall Q$, and assume a uniform prior over reward parameters $\omega$, i.e., $P(\omega)$ is constant for any $\norm{\omega}_2\leq1$. We use Metropolis-Hastings algorithm for sampling the set $\Omega$ from belief distribution over $\omega$}. We start by describing the simulation domains and the user study environment, and introducing the human choice models we employed for preference-based learning. Each subsequent subsection presents a set of experiments and tests the relevant hypotheses.

\begin{figure}[t]
\includegraphics[width=\columnwidth]{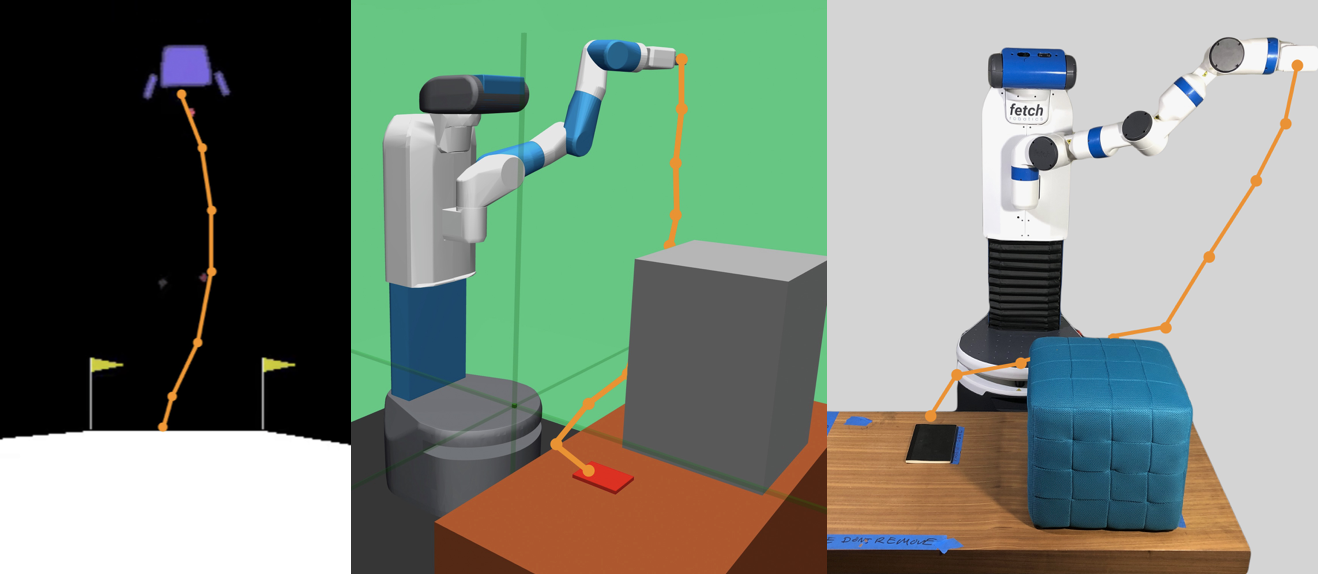}
\centering
\caption{Views from simulation domains, with a demonstration in orange: \textbf{(a)} Lunar Lander, \textbf{(b)} Fetch (simulated), \textbf{(c)} Fetch (physical). }
\label{fig:rss19_domains}
\end{figure}

\smallskip

\noindent \textbf{Simulation Domains.} In each experiment, we use a subset of the following domains, shown in Figures \ref{fig:corl19_domains} and \ref{fig:rss19_domains}, as well as a linear dynamical system:

\textit{Linear Dynamical System (LDS)}: We use a linear dynamical system with six dimensional state and three dimensional action spaces. State values are directly used as state features without any transformation.

\textit{Driver}: We use a 2D driving simulator \citep{sadigh2016planning}, where the agent has to safely drive down a highway. The trajectory features correspond to the distance of the agent from the center of the lane, its speed, heading angle, and minimum distance to other vehicles during the trajectory (white in Figure \ref{fig:corl19_domains}~(top)).

\textit{Tosser}: We use a Tosser robot simulation built in MuJoCo \citep{todorov2012mujoco} that tosses a capsule-shaped object into baskets. The trajectory features are the maximum horizontal distance forward traveled by the object, the maximum altitude of the object, the number of flips the object does, and the object's final distance to the closest basket. 

\textit{Lunar Lander}: We use the continuous LunarLander environment from OpenAI Gym \citep{brockman2016openai}, where the lander has to safely reach the landing pad. The trajectory features correspond to the lander's average distance from the landing pad, its angle, its velocity, and its final distance to the landing pad. 

\textit{Fetch}: Inspired by \cite{bajcsy2017learning}, we use a modification of the Fetch Reach environment from OpenAI Gym (built on top of MuJoCo), where the robot has to reach a goal with its arm, while keeping its arm as low-down as possible (see Figure~\ref{fig:rss19_domains}). The trajectory features correspond to the robot gripper's average distance to the goal, its average height from the table, and its average distance to a box obstacle in the domain.

For our user studies, we employ a version of the Fetch environment with the physical Fetch robot (see Figure~\ref{fig:rss19_domains}) \citep{wise2016fetch}.

\smallskip

\noindent \textbf{Human Choice Models.} We require a probabilistic model for the human's choice $q$ in a query $Q$ conditioned on their reward parameters $\omega$. Below, we discuss two specific models that we use in our experiments.

\textit{Standard Model Using Strict Preference Queries.} Previous work demonstrated the importance of modeling imperfect human responses \citep{kulesza2014structured}. We model a noisily optimal human as selecting $\xi_k$ from a \emph{strict} preference query $Q$ by
\begin{align}
P(q=\xi_k\mid Q,\omega) &= \frac{\exp(R(\xi_k))}{\sum_{\xi\in Q}\exp(R(\xi))}, 
\label{eq:noisily_optimal}
\end{align}
where we call the query strict because the human is required to select one of the trajectories. This model, backed by neuroscience and psychology \citep{daw2006cortical,luce2012individual,ben1985discrete,lucas2009rational}, is routinely used  \citep{biyik2019green, guo2010real, viappiani2010optimal}.

\textit{Extended Model Using Weak Preference Queries.} We generalize this preference model to include an ``About Equal" option for queries between two trajectories.
We denote this option by $\Upsilon$ and define a \emph{weak preference query} $Q^+ := Q\cup\{\Upsilon\}$ when $K=2$.

Building on prior work by \cite{krishnan1977incorporating}, we incorporate the information from the ``About Equal" option by introducing a minimum perceivable difference parameter $\delta \geq 0$, and defining:
\begin{align}
P&(q=\Upsilon \mid Q^+,\omega) = \nonumber\\
&\left(\exp(2\delta) - 1\right)P(q=\xi_1\mid  Q^+,\omega)P(q=\xi_2 \mid  Q^+,\omega)\:, \nonumber\\
P&(q=\xi_k\mid Q^+,\omega) = \nonumber\\
&\frac{1}{1\!+\!\exp(\delta \!+\! R(\xi_{k'}) \!-\! R(\xi_k))}, \{\xi_k,\xi_{k'}\} \!=\! Q^+\!\setminus\!\{\Upsilon\}.
\label{eq:weak_noisily_optimal}
\end{align}
Notice that Equation~\eqref{eq:weak_noisily_optimal} reduces to Equation~\eqref{eq:noisily_optimal} when $\delta=0$; in which case we model the human as always perceiving the difference in options.
All derivations in earlier sections hold with weak preference queries. In particular, we include a discussion of extending our formulation to the case where $\delta$ is user-specific and unknown in the Appendix. The additional parameter causes no trouble in practice. For all our experiments, we set $K=2$, and $\delta=1$ (whenever relevant).

We note that there are alternative choice models compatible with our framework for weak preferences (e.g., \citep{holladay2016active}).
Additionally, one may generalize the weak preference queries to $K>2$, though it complicates the choice model as the user must specify which of the trajectories create uncertainty.

\smallskip

\noindent \textbf{Evaluation Metric.} To judge convergence of inferred reward parameters to true parameters in simulations, we adopt the \emph{alignment metric} from \cite{sadigh2017active}: 
\begin{align}
m\!=\!\frac{1}{M}\sum_{\bar{\omega}\in\Omega}\frac{\omega^*\cdot\bar{\omega}}{\norm{\omega^*}_2\norm{\bar{\omega}}_2},
\end{align}
where $\omega^*$ is the true reward parameters.

We are now ready to present our five sets of experiments each of which demonstrates a different aspect of the proposed DemPref framework:
\begin{enumerate}[nosep]
    \item The utility of initializing with demonstrations,
    \item The advantages preference queries provide over using only demonstrations,
    \item The advantages of information gain formulation over volume removal,
    \item The order of demonstrations and preferences, and
    \item Optimal stopping condition under the information gain objective.
\end{enumerate}

\subsection{Initializing with Demonstrations}
We first investigate whether initializing the learning framework with user demonstrations is helpful. Specifically, we test the following hypotheses:

\vspace{3px}
\noindent\textbf{H1.} \emph{DemPref accelerates learning by initializing the prior belief $b^0$ using user demonstrations.}
\vspace{3px}

\noindent\textbf{H2.} \emph{The convergence of DemPref improves with the number of demonstrations used to initialize the algorithm.}
\vspace{3px}

To test these two claims, we perform simulation experiments in Driver, Lunar Lander and Fetch environments. For each environment, we simulate a human user with hand-tuned reward function parameters $\omega$, which gives reasonable performance. We generate demonstrations by applying model predictive control (MPC) to solve: $\max_\Lambda \omega^*\cdot\Phi(\Lambda)$. After initializing the belief with varying number of such demonstrations ($\abs{\mathcal{D}}\in\{0,1,3\}$), the simulated users in each environment respond to $25$ strict preference queries according to Equation~\eqref{eq:noisily_optimal}, each of which is actively synthesized with the volume removal optimization. We repeat the same procedure for $8$ times to obtain confidence bounds.

\begin{figure*}[h]
\includegraphics[width=0.85\textwidth]{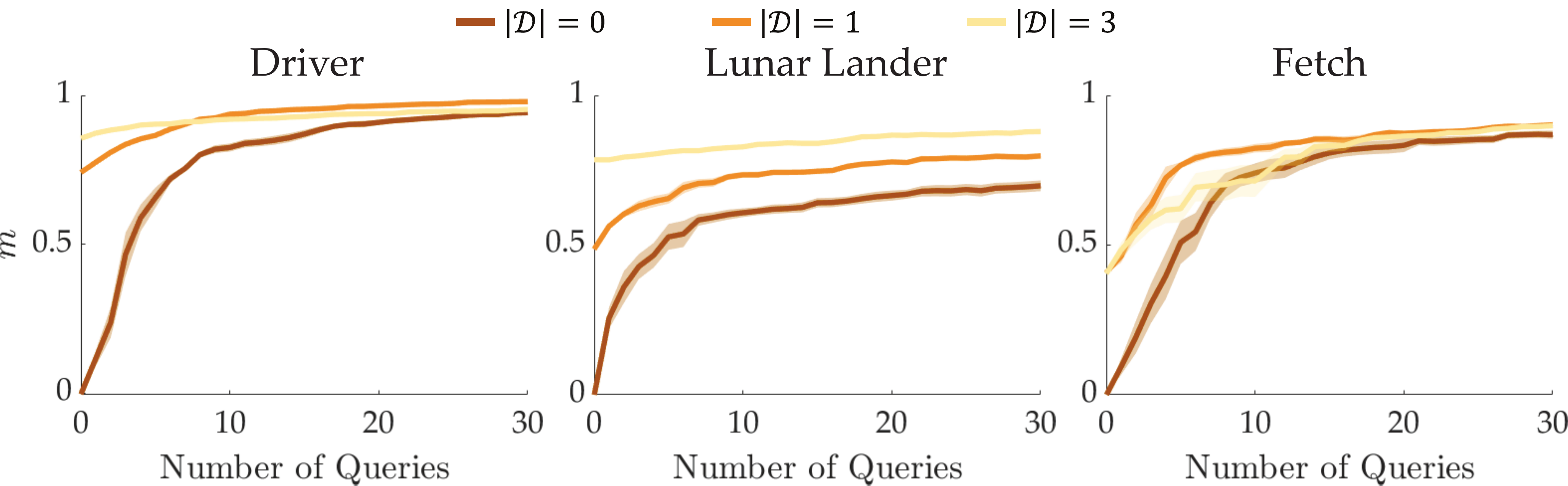}
\centering
\caption{The results of our first experiment, investigating whether initializing with demonstrations improves the learning rate of the algorithm, on three domains. On the Driver, Lunar Lander, and Fetch environments, initializing with one demonstration improved the rate of convergence significantly.}
\label{fig:varying_D}
\end{figure*}

The results of the experiment are presented in Figure \ref{fig:varying_D}. On all three environments, initializing with demonstrations improves the convergence rate of the preference-based algorithm significantly; to match the $m$ value attained by DemPref with only one demonstration in $10$ preference queries, it takes the pure preference-based algorithm, i.e., without any demonstrations, $30$ queries on Driver, $35$ queries on Lander, and $20$ queries on Fetch. These results provide strong evidence in favor of \textbf{H1}.

The results regarding \textbf{H2} are more complicated. Initializing with three instead of one demonstration improves convergence significantly only on the Driver and Lunar Lander domains. (The improvement on Driver is only at the early stages of the algorithm, when fewer than 10 preferences are used.) However, on the Fetch domain, initializing with three instead of one demonstration hurts the performance of the algorithm. (Although, we do note that the results from using three demonstrations are still an improvement over the results from not initializing with demonstrations). This is unsurprising. It is much harder to provide good demonstrations on the Fetch environment than on the Driver or Lunar Lander environments, and therefore the demonstrations are of lower quality. Using more demonstrations when they are of lower quality leads to the prior being more concentrated further away from the true reward function, and can cause the preference-based learning algorithm to slow down.

In practice, we find that using a single demonstration to initialize the algorithm leads to reliable improvements in convergence, regardless of the complexity of the domain.

\subsection{DemPref vs IRL}
Next, we analyze if preference elicitation improves learning performance. To do that, we conduct a within-subjects user study where we compare our DemPref algorithm with Bayesian IRL \citep{ramachandran2007bayesian}. The hypotheses we are testing are:

\vspace{3px}
\noindent\textbf{H3.} \emph{The robot which uses the reward function learned by DemPref will be more successful at the task (as evaluated by the users) than the IRL counterpart.}
\vspace{3px}

\noindent\textbf{H4.} \emph{Participants will prefer to use the DemPref framework as opposed to the IRL framework.}
\vspace{3px}

For these evaluations, we use the Fetch domain with the physical Fetch robot. Participants were told that their goal was to get the robot's end-effector as close as possible to the goal, while (1) avoiding collisions with the block obstacle and (2) keeping the robot's end-effector low to the ground (so as to avoid, for example, knocking over objects around it). Participants provided demonstrations via teleoperation (using end-effector control) on a keyboard interface; each user was given some time to familiarize themselves with the teleoperation system before beginning the experiment.

Participants trained the robot using two different systems. (1) IRL: Bayesian IRL  with 5 demonstrations. (2) DemPref: our DemPref framework (with the volume removal objective) with 1 demonstration and 15 proactive preference queries\footnote{The number of demonstrations and preferences used in each system were chosen such that a simulated agent achieves similar convergence to the true reward on both systems.}. We counter-balanced across which system was used first, to minimize the impact of familiarity bias with our teleoperation system.

\begin{figure*}[t]
\includegraphics[width=\textwidth]{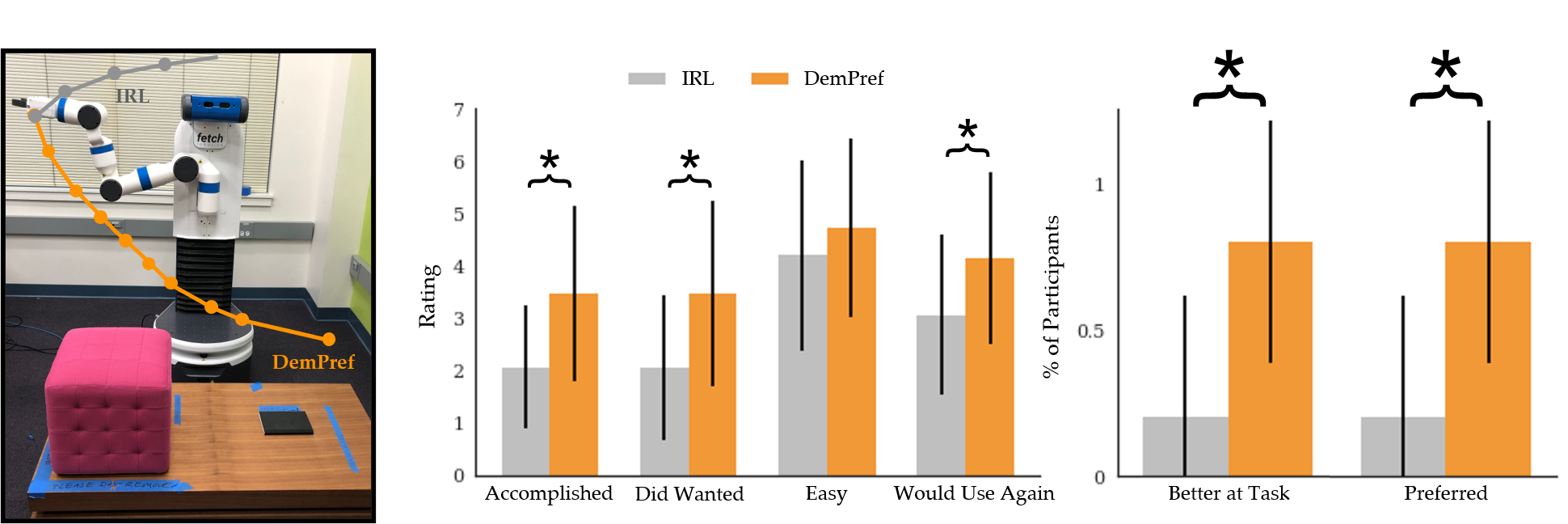}
\centering
\caption{(Left) Our testing domain, with two trajectories generated according to the reward functions learned by IRL and DemPref from a specific user in our study. (Right) The results of our usability study -- the error bars correspond to standard deviation and significant results are marked with an asterisk. We find that users rated the robot trained with DemPref as significantly better at accomplishing the task and preferred to use our method for training the robot significantly more than they did IRL. However, we did not find evidence to suggest that users found our method easier to use than standard IRL.}
\label{fig:rss_user_study}
\vspace{-10px}
\end{figure*}

After training, the robot was trained in simulation using Proximal Policy Optimization (PPO) with the reward function learned from each system \citep{schulman2017proximal}. To ensure that the robot was not simply overfitting to the training domain, we used different variants of the domain for training and testing the robot. We used two different test domains (and counter-balanced across them) to increase the robustness of our results against the specific testing domain. Figure \ref{fig:rss_user_study}~(left) illustrates one of our testing domains.
We rolled out three trajectories in the test domains for each algorithm on the physical Fetch. After observing each set of trajectories, the users were asked to rate the following statements on a 7-point Likert scale:
\begin{enumerate}[nosep]
    \item The robot accomplished the task well. (Accomplished)
    \item The robot did what I wanted. (Did Wanted)
    \item It was easy to train the robot with this system. (Easy)
    \item I would want to use this system to train a robot in the future. (Would Use Again)
\end{enumerate}
They were also asked two comparison questions:
\begin{enumerate}[nosep]
    \item Which robot accomplished the task better? (Better at Task)
    \item Which system would you prefer to use if you had to train a robot to accomplish a similar task? (Preferred)
\end{enumerate}
They were finally asked for general comments.

For this user study, we recruited 15 participants (11 male, 4 female), six of whom had prior experience in robotics but none of whom had any prior exposure to our system.

We present our results in Figure~\ref{fig:rss_user_study}~(right). When asked which robot accomplished the task better, users preferred the DemPref system by a significant margin ($p < 0.05$, Wilcoxon paired signed-rank test); similarly, when asked which system they would prefer to use in the future if they had to train the robot, users preferred the DemPref system by a significant margin ($p < 0.05$). This provides strong evidence in favor of both \textbf{H3} and \textbf{H4}. 

As expected, many users struggled to teleoperate the robot. Several users made explicit note of this fact in their comments: ``I had a hard time controlling the robot", ``I found the [IRL system] difficult as someone who [is not] kinetically gifted!", ``Would be nice if the controller for the [robot] was easier to use." Given that the robot that employs IRL was only trained on these demonstrations, it is perhaps unsurprising that DemPref outperforms IRL on the task.

We were however surprised by the extent to which the IRL-powered robot fared poorly: in many cases, it did not even attempt to reach for the goal. Upon further investigation, we discovered that IRL was prone to, in essence, ``overfitting" to the training domain. In several cases, IRL had overweighted the users' preference for obstacle avoidance. This proved to be an issue in one of our test domains where the obstacle is closer to the robot than in the training domain. Here, the robot does not even try to reach for the goal since the loss in value (as measured by the learned reward function) from going near the obstacle is greater than the gain in value from reaching for the goal. Figure \ref{fig:rss_user_study}~(left) shows this test domain and illustrates, for a specific user, a trajectory generated according to reward function learned by each of IRL and DemPref.

While we expect that IRL would overcome these issues with more careful feature engineering and increased diversity of the training domains, it is worth noting DemPref was affected much less by these issues. These results suggest preference-based learning methods may be more robust to poor feature engineering and a lack of training diversity than IRL; however, a rigorous evaluation of these claims is beyond the scope of this paper. 

It is interesting that despite the challenges that users faced with teleoperating the robot, they did not rate the DemPref system as being ``easier" to use than the IRL system ($p = 0.297$). Several users specifically referred to the time it took to generate each query ($\sim$45 seconds) as negatively impacting their experience with the DemPref system: ``I wish it was faster to generate the preference [queries]", ``The [DemPref system] will be even better if time cost is less." Additionally, one user expressed difficulty in evaluating the preference queries themselves, commenting ``It was tricky to understand/infer what the preferences were [asking]. Would be nice to communicate that somehow to the user (e.g. which [trajectory] avoids collision better)!", which highlights the fact that volume removal formulation may generate queries that are extremely difficult for the humans. Hence, we analyze in the next subsection how information gain objective improves the experience for the users.

\subsection{Information Gain vs Volume Removal}
To investigate the performance and user-friendliness of the information gain and volume removal methods for preference-based learning, we conduct experiments with simulated users in LDS, Driver, Tosser and Fetch environments; and real user studies in Driver, Tosser and Fetch (with the physical robot). We are particularly interested in the following three hypotheses:

\vspace{3px}
\noindent\textbf{H5.} \emph{Information gain formulation outperforms volume removal in terms of data-efficiency.}
\vspace{3px}

\noindent\textbf{H6.} \emph{Information gain queries are easier and more intuitive for the human than those from volume removal.}
\vspace{3px}

\noindent\textbf{H7.} \emph{A user's preference aligns best with reward parameters learned via information gain.} 
\vspace{3px}

To enable faster computation, we discretized the search space of the optimization problems by generating $500,000$ random queries and precomputing their trajectory features. Each call to an optimization problem then performs a loop over this discrete set.

\begin{figure*}[t]
	\centering
	\includegraphics[width=\textwidth]{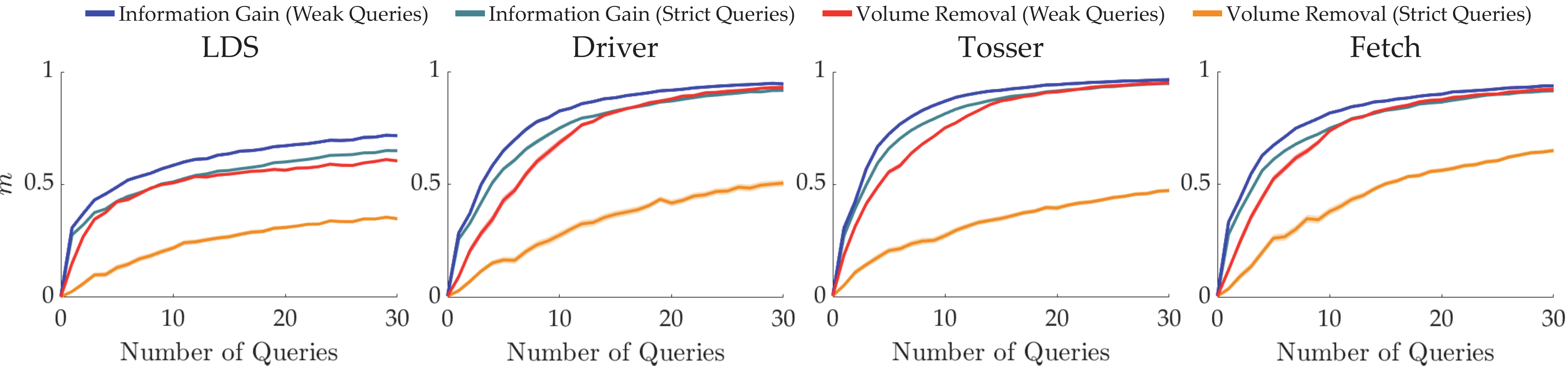}
	\vspace{-15px}
	\caption{Alignment values are plotted (mean $\pm$ standard error) to compare information gain and volume removal formulations. Standard errors are so small that they are mostly invisible in the plots. Dashed lines show the weak preference query variants. Information gain provides a significant increase in learning rate in all cases. While weak preference queries lead to a large amount of improvement under volume removal, information gain formulation is still superior in terms of the convergence rate.}
	\label{fig:simulation_results_m}
\end{figure*}

In simulation experiments, we learn the randomly generated reward functions via both strict and weak preference queries where the ``About Equal" option is absent and present, respectively. We repeat each experiment $100$ times to obtain confidence bounds. Figure~\ref{fig:simulation_results_m} shows the alignment value against query number for the $4$ different tasks. Even though the ``About Equal" option improves the performance of volume removal by preventing the trivial query, $Q=\{\xi_A,\xi_A,\dots\}$, from being a global optimum, information gain gives a significant improvement on the learning rate both with and without the ``About Equal" option in all environments\footnote{See the Appendix for results without query space discretization.}. These results strongly support \textbf{H5}.

\begin{figure*}[t]
	\centering
	\includegraphics[width=\textwidth]{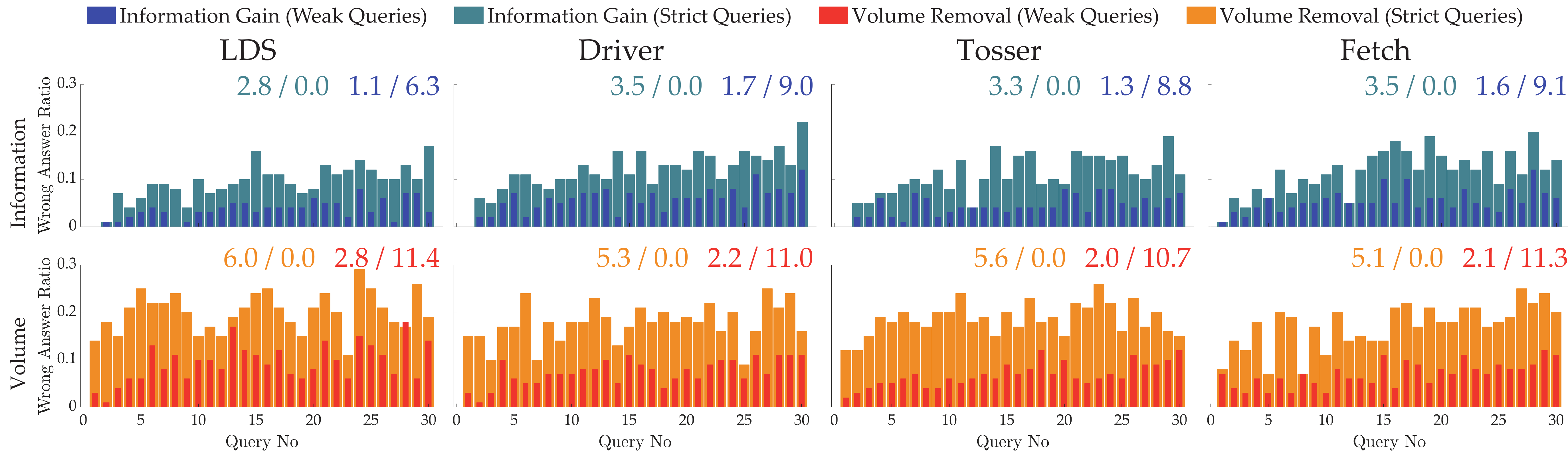}
	\vspace{-15px}
	\caption{Wrong answer ratios on different queries are shown. The numbers at top show the average number of wrong responses and ``About Equal" choices, respectively, for both strict and weak queries. Information gain formulation yields smaller numbers of wrong and ``About Equal" answers, especially in the early stages.}
	\label{fig:simulation_results_wrong_responses}
	\vspace{-5px}
\end{figure*}

The numbers given within Figure~\ref{fig:simulation_results_wrong_responses} count the wrong answers and ``About Equal" choices made by the simulated users. The information gain formulation significantly improves over volume removal. Moreover, weak preference queries consistently decrease the number of wrong answers, which can be one reason why it performs better than strict queries\footnote{Another possible explanation is the information acquired by the ``About Equal" responses. We analyze this in the Appendix by comparing the results with what would happen if this information was discarded.}. Figure~\ref{fig:simulation_results_wrong_responses} also shows when the wrong responses are given. While wrong answer ratios are higher with volume removal formulation, it can be seen that information gain reduces wrong answers especially in early queries, which leads to faster learning. These results support \textbf{H6}.

In the user studies for this part, we used Driver and Tosser environments in simulation and the Fetch environment with the physical robot. We began by asking participants to rank a set of features (described in plain language) to encourage each user to be consistent in their preferences. Subsequently, we queried each participant with a sequence of 30 questions generated actively; 15 from volume removal and 15 via information gain.
We prevent bias by randomizing the sequence of questions for each user and experiment: the user does not know which algorithm generates a question.

Participants responded to a 7-point Likert scale survey after each question:
\begin{enumerate}[nosep]
    \item It was easy to choose between the trajectories that the robot showed me.
\end{enumerate}
They were also asked the Yes/No question:
\begin{enumerate}[nosep]
    \item Can you tell the difference between the options presented?
\end{enumerate}

In concluding the Tosser and Driver experiments, we showed participants two trajectories: one optimized using reward parameters from information gain (trajectory A) and one optimized using reward parameters from volume removal (trajectory B)\footnote{We excluded Fetch for this question to avoid prohibitive trajectory optimization (due to large action space).}.
Participants responded to a 7-point Likert scale survey: \begin{enumerate}[nosep]
    \item Trajectory A better aligns with my preferences than trajectory B.
\end{enumerate}

We recruited $15$ participants (8 female, 7 male) for the simulations (Driver and Tosser) and $12$ for the Fetch (6 female, 6 male). We used strict preference queries. A video demonstration of these user studies is available at \url{http://youtu.be/JIs43cO\_g18}.

\begin{figure*}[t]
	\centering
	\includegraphics[width=0.85\textwidth]{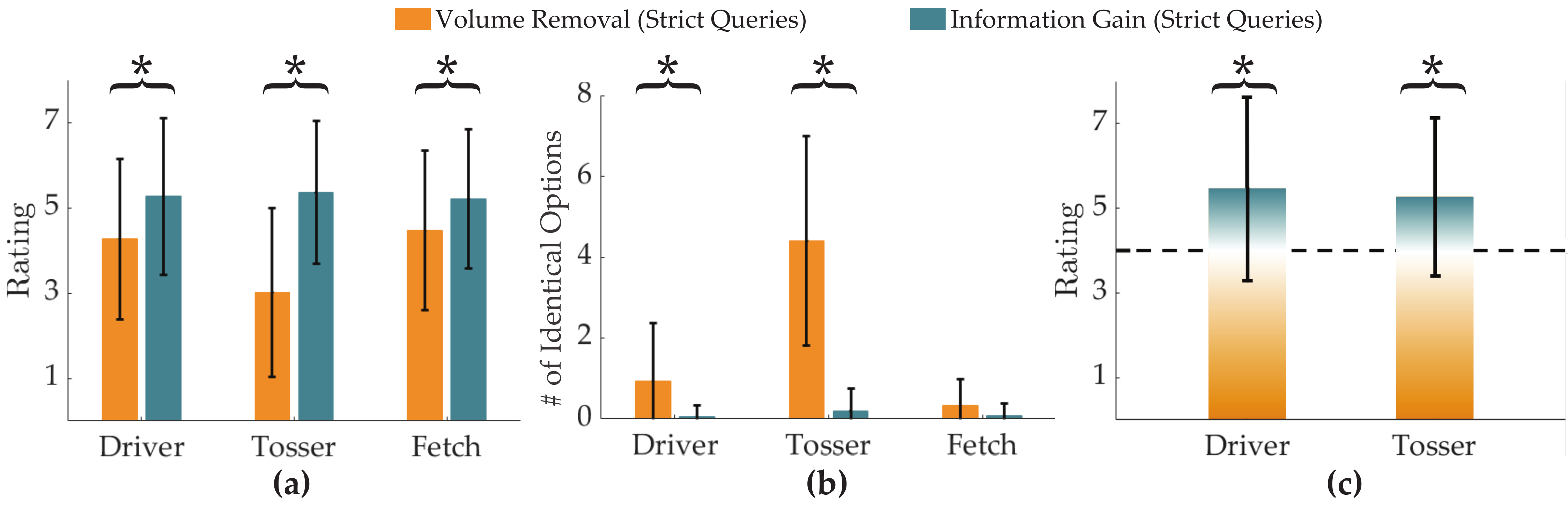}
	\vspace{-8px}
	\caption{User study results. Error bars show std. Asterisks show statistical significance. \textbf{(a)} Easiness survey results averaged over all queries and users. Queries generated using the information gain method are rated significantly easier by the users than the volume removal queries. \textbf{(b)} The number of identical options in the experiments averaged over all users. In Driver and Tosser, users indicated significantly less indistinguishable queries with information gain compared to volume removal.  \textbf{(c)} Final preferences averaged over the users. $7$ means the user strongly prefers the optimized trajectory w.r.t. the learned reward by the information gain formulation, and $1$ is the volume removal. Dashed line represents indifference between two methods.  Users significantly prefer the robot who learned using the information gain method for active query generation.}
	\label{fig:corl_user_studies}
	\vspace{-10px}
\end{figure*}

Figure~\ref{fig:corl_user_studies}~(a) shows the results of the easiness surveys.
In all environments, users found information gain queries easier: the results are statistically significant ($p<0.005$, two-sample $t$-test). Figure~\ref{fig:corl_user_studies}~(b) shows the average number of times the users stated they cannot distinguish the options presented. The volume removal formulation yields several queries that are indistinguishable to the users while the information gain avoids this issue. The difference is significant for Driver ($p<0.05$, paired-sample $t$-test) and Tosser ($p<0.005$). Taken together, these results support \textbf{H6}.

Figure~\ref{fig:corl_user_studies}~(c) shows the results of the survey the participants completed at the end of experiment. Users significantly preferred the information gain trajectory over that of volume removal in both environments ($p<0.05$, one-sample $t$-test), supporting \textbf{H7}.

\subsection{The Order of Information Sources}
Having seen information gain provides a significant boost in the learning rate, we checked whether the passively collected demonstrations or the actively queried preferences should be given to the model first. Specifically, we tested:

\vspace{3px}
\noindent\textbf{H8.} \emph{If passively collected demonstrations are used before the actively collected preference query responses, the learning becomes faster.}
\vspace{3px}

While Theorem~\ref{thm:order_of_dempref} asserts that we should first initialize DemPref via demonstrations, we performed simulation experiments to check this notion in practice. Using LDS, Driver, Tosser and Fetch, we ran three sets of experiments where we adopted weak preference queries: (i) We initialize the belief with a single demonstration and then query the simulated user with $15$ preference questions, (ii) We first query the simulated user with $15$ preference questions and we add the demonstration to the belief independently after each question, and (iii) We completely ignore the demonstration and use only $15$ preference queries. The reason why we chose to have only a single demonstration is because having more more demonstrations tends to increase the alignment value $m$ for both (i) and (ii), thereby making the difference between the methods' performances very small. We ran each set of experiment $100$ times with different, randomly sampled, true reward functions. We again used the same data set of $500,000$ queries for query generation. We also used the trajectory that gives the highest reward to the simulated user out of this data set as the demonstration in the first two sets of experiments. Since the demonstration is not subject to noises or biases due to the control of human users, we set $\beta^D=0.2$.

\begin{figure*}[th]
	\centering
	\includegraphics[width=\textwidth]{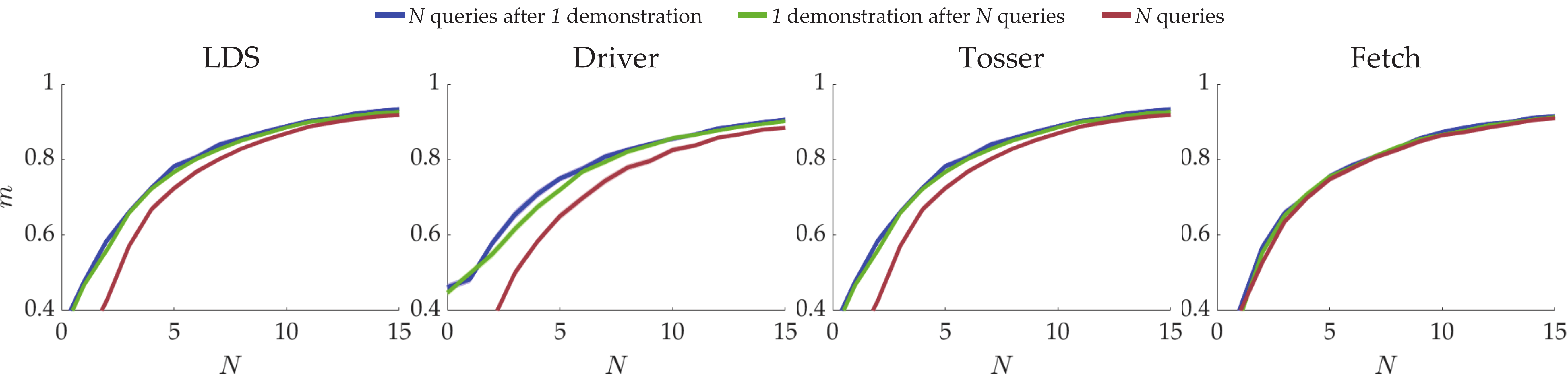}
	\vspace{-15px}
	\caption{Simulation results for the order of demonstrations and preference queries. Alignment values are plotted (mean$\pm$s.e.). It is consistently better to first utilize the passively collected demonstrations rather than actively generated preference queries. The differences in the alignment value is especially small in the Fetch simulations, which might be due to the fact that it is a simpler environment in terms of the number of trajectory features.}
	\label{fig:demfirst_vs_demsecond}
	\vspace{-5px}
\end{figure*}

Figure~\ref{fig:demfirst_vs_demsecond} shows the alignment value against the number of queries. The last set of experiments has significantly lower alignment values than the first two sets especially when the number of preference queries is small. This indicates the demonstration has carried an important amount of information. Comparing the first two sets of experiments, the differences in the alignment values are small. However, the values are consistently higher when the demonstrations are used to initialize the belief distribution. This supports \textbf{H8} and numerically validates Theorem~\ref{thm:order_of_dempref}.

\subsection{Optimal Stopping}
Finally, we experimented our optimal stopping extension for information gain based active querying algorithm in LDS, Driver, Tosser and Fetch environments with simulated users. Again adopting query discretization, we tested:

\vspace{3px}
\noindent\textbf{H9.} \emph{Optimal stopping enables cost-efficient reward learning under various costs.}
\vspace{3px}

\begin{figure*}[th]
	\centering
	\includegraphics[width=\textwidth]{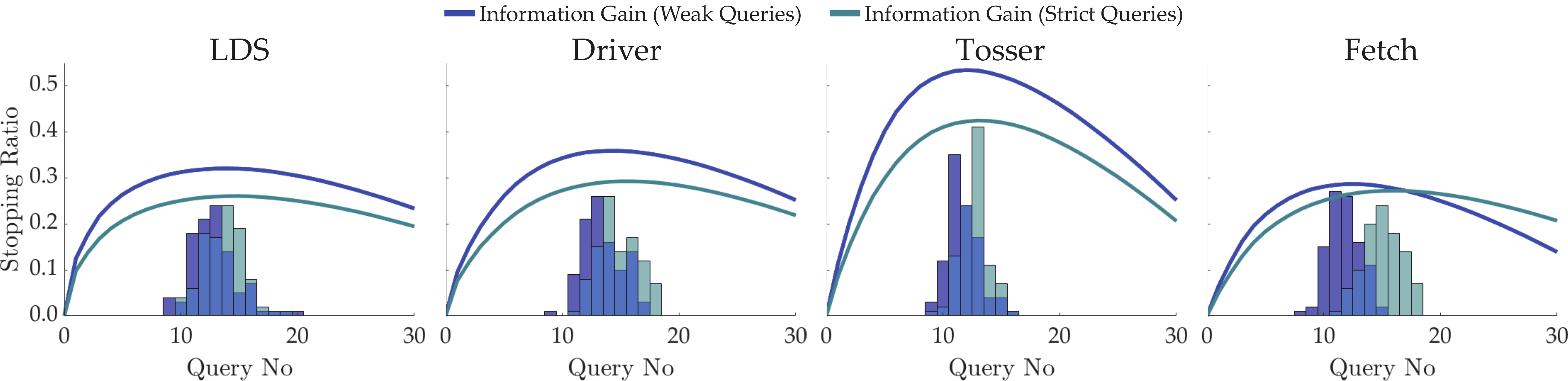}
	\vspace{-15px}
	\caption{Simulation results for optimal stopping. Line plots show cumulative active learning rewards (cumulative difference between the information gain values and the query costs), averaged over $100$ test runs and scaled for visualization. Histograms show when optimal stopping condition is satisfied, which aligns with the desired cumulative rewards.}
	\label{fig:optimal_stopping_query_dependent}
	\vspace{-8px}
\end{figure*}

As the query cost, we employed a cost function to improve interpretability of queries, which may have the associated benefit of making learning more efficient \citep{bajcsy2018learning}. We defined a cost function:
\begin{align*}
c(Q) = \epsilon-\abs{\Psi_{i^*}} + \max_{j\in\{1,\dots\}\setminus\{i^*\}}\abs{\Psi_j},\: i^*=\argmax_i{\abs{\Psi_i}},
\end{align*}
where $Q=\{\xi_1,\xi_2\}$ and $\Psi=\Phi(\xi_1)-\Phi(\xi_2)$.
This cost favors queries in which the difference in one feature is larger than that between all other features. Such a query may prove more interpretable.
We first simulate $100$ random users and tune $\epsilon$ accordingly: For each simulated user, we record the $\epsilon$ value that makes the objective zero in the $i^{\textrm{th}}$ query (for smallest $i$) such that $m_i, m_{i-1}, m_{i-2} \in [x,x+0.02]$ for some $x$. We then use the average of these $\epsilon$ values for our tests with $100$ different random users. Figure~\ref{fig:optimal_stopping_query_dependent} shows the results\footnote{We found similar results with query-independent costs minimizing the number of queries. See Appendix.}. Optimal stopping rule enables terminating the process with near-optimal cumulative active learning rewards (the cumulative difference between the information gain and the cost as in Equation~\eqref{eq:EX1}) in all environments, which supports \textbf{H9}.

\section{Conclusion}

\noindent \textbf{Summary.} In this paper, we developed a framework that utilizes multiple sources of information to learn the preference reward functions of the human users, demonstrations and preference queries in particular. In addition to proving that active information sources must be used after the passive sources for more efficient learning, we highlighted the important problems associated with the state-of-the-art active preference query generation method. The alternative technique we proposed, information gain maximization, has proven to not suffer from those problems while being at least tractable. Moreover, our proposed technique allows the designers to incorporate an optimal stopping condition, which improves the usefulness of the entire framework. We performed a large number of simulation experiments and user studies on a physical robot whose results have demonstrated the superiority of our proposed method over various state-of-the-art alternatives both in terms of performance and user preferences, as well as the validity of the optimal stopping criterion and the order of information sources.

\smallskip

\noindent \textbf{Limitations and Future Work.} In this work, we focused on two common sources of information: demonstrations and preference questions. While we showed that active information sources must be used after the passive ones, our work is limited in terms of the number of information sources considered: once multiple active information sources are available, their optimal order may depend on various factors, such as how informative or how costly they are. Besides, the information sources we adopt in this paper can be further extended. For example, ``About Equal" option in preference queries allowed users to give more expressive feedbacks. One could also think of other forms of responses, such as ``I (dis)like both" or ``I like this more than the other by this much", in which cases a better human choice model would be needed. Secondly, we considered a standard reward model in IRL which is linear in some trajectory features. While this might be expressive enough with well-designed features, it might be necessary in some cases to account for nonlinearities. Even though we recently developed an active preference query generation method for rewards modeled using Gaussian processes to tackle this issue \citep{biyik2020active}, incorporating demonstrations to this framework to attain a nonlinear version of DemPref might be nontrivial. 
Third, our work is limited in terms of the number of baselines used for comparison. While, to the best of our knowledge, prior works do not incorporate demonstrations and preference queries, each component could be individually tested against other baselines (e.g., \cite{brown2019deep,michini2012bayesian,park2020inferring}), which might inform a better DemPref algorithm. Recent research which builds on our approach highlights that preference questions generated with information gain --- although easy to answer --- may mislead the human into thinking the robot does not understand what it has actually learned \citep{habibian2021here}.
Finally, our results showed having too many demonstrations, which are imprecise sources of information, might harm the performance of DemPref. An interesting future research direction is then to investigate the optimal number of demonstrations, and to decide when and which demonstrations are helpful.

\section*{Acknowledgments}
This work is supported by FLI grant RFP2-000, and NSF Awards \#1849952 and \#1941722.

\bibliographystyle{SageH}
\bibliography{citations}

\section{Appendix}

In the appendix, we present
\begin{enumerate}[nosep]
    \item The derivation of the information gain formulation,
    \item The proof of Theorem~\ref{thm:optimal_stopping},
    \item The extension of the information gain formulation to the case with user-specific and unknown $\delta$ under weak preference queries,
    \item The comparison of information gain and volume removal formulations without query space discretization, i.e., with continuous trajectory optimization,
    \item The experimental analysis of the effect of the information coming from ``About Equal" responses  in weak preference queries, and
    \item Optimal stopping under query-independent costs.
\end{enumerate}

\subsection{Derivation of Information Gain Solution}
\label{sec:info_gain_derivation}

We first present the full derivation of Equation~\eqref{eq:IG2},
\begin{align*}
Q^*_i = \argmax_{Q_i=\{\Lambda_1,\dots,\Lambda_K\}} I(q_i ; \omega \mid  Q_i, b^i)\textcolor{red}{\:.}\\
\end{align*}
We first write the mutual information as the difference between two entropy terms:
\begin{align*}
I&(q_i ; \omega \mid  Q_i, b^i)\\
&= H(\omega \mid  Q_i, b^i) - \mathbb{E}_{q_i\mid Q_i,b^i}\left[H(\omega \mid  q_i,Q_i,b^i)\right]\textcolor{red}{\:.}
\end{align*}
Next, we expand the entropy expressions and use $P(q_i\mid Q_i,b^i)P(\omega\mid q_i,Q_i,b^i)=P(\omega,q_i\mid Q_i,b^i)$ to combine the expectations for the second term to get:
\begin{align*}
H&(\omega \mid  Q_i, b^i) - \mathbb{E}_{q_i\mid Q_i,b^i}\left[H(\omega \mid  q_i,Q_i,b^i)\right]\\
&= -\mathbb{E}_{\omega\mid Q_i,b^i}\left[\log_2 P(\omega\mid Q_i,b^i)\right] + \\&\qquad \mathbb{E}_{\omega,q_i\mid Q_i,b^i}\left[\log_2\left(P(\omega \mid  q_i,Q_i,b^i)\right)\right]\textcolor{red}{\:.}
\end{align*}
Since the first term is independent from $q_i$, we can write this expression as
\begin{align*}
&\mathbb{E}_{\omega,q_i\mid Q_i,b^i}\left[\log_2 P(\omega \mid  q_i,Q_i,b^i)\!-\!\log_2 P(\omega\mid Q_i,b^i)\right]\\
&= \mathbb{E}_{\omega, q_i\mid Q_i,b^i}\left[\log_2 P(q_i \mid  Q_i,b^i,\omega)\!-\!\log_2 P(q_i\mid Q_i,b^i)\right]\\
&= \mathbb{E}_{\omega, q_i\mid Q_i,b^i}\Bigg[\log_2 P(q_i \mid  Q_i,\omega) - \\&\qquad\log_2\left(\int P(q_i\mid Q_i,\omega')P(\omega' \mid  Q_i,b^i)d\omega'\right)\Bigg] \;,
\end{align*}
where the integral is taken over all possible values of $\omega$.

Having $\Omega$ as a set of $M$ samples drawn from the prior $b^i$,
\begin{align*}
I&(q_i ; \omega \mid  Q_i)\\
&\asymeq \mathbb{E}_{\omega, q_i\mid Q_i}\Bigg[\log_2 P(q_i \mid  Q_i,\omega) - \\&\qquad\log_2\left(\frac1M \sum_{\omega'\in\Omega} P(q_i\mid Q_i,\omega')\right)\Bigg]\\
&= \mathbb{E}_{\omega, q_i\mid Q_i}\left[\log_2\frac{M\cdot P(q_i \mid  Q_i,\omega)}{ \sum_{\omega'\in\Omega} P(q_i\mid Q_i,\omega')}\right]\\
&= \mathbb{E}_{\omega\mid Q_i}\!\left[\mathbb{E}_{q_i\mid Q_i,\omega}\left[\log_2\frac{M\cdot P(q_i \mid  Q_i,\omega)}{ \sum_{\omega'\in\Omega} P(q_i\!\mid\! Q_i,\omega')}\right]\right]\\
&= \mathbb{E}_{\omega\mid Q_i}\Bigg[\!\sum_{q_i\!\in Q_i}\! P(q_i \!\mid\! Q_i,\!\omega) \log_2\frac{M\cdot P(q_i \mid  Q_i,\omega)}{ \sum_{\omega'\!\in\Omega} P(q_i \!\mid\! Q_i,\!\omega')}\!\Bigg]\\
&\asymeq \frac1M\! \sum_{q_i\!\in Q_i}\!\sum_{\bar{\omega}\!\in\Omega}\! P(q_i \!\mid\! Q_i,\!\bar{\omega}) \log_2\frac{M\cdot P(q_i \mid  Q_i,\bar{\omega})}{ \sum_{\omega'\!\in\Omega} P(q_i\!\mid\! Q_i,\!\omega')} \;,
\end{align*}
where, in the last step, we use the sampled $\omega$'s to compute the expectation over $\omega\mid Q_i$. This completes the derivation.

\subsection{Proof of Theorem 3}
\setcounter{theorem}{2}
\begin{theorem}
	Terminating the algorithm is optimal if and only if global solution to \eqref{eq:EX1} is negative.
	\begin{proof}
	 	We need to show if the global optimum is negative, then any longer-horizon optimization will also give negative reward (difference between information gain and the cost) in expectation. Let $Q^*_i$ denote the global optimizer. For any $k\geq0$,
	 	\begin{align*}
	 	I&(q_i,\dots,q_{i+k} ; \omega \mid  Q_i,\dots,Q_{i+k}) - \sum_{j=0}^{k}c(Q_{i+j}) \\
	 	&= I(q_i ; \omega \mid  Q_i) + \dots + \\&\quad I(q_{i+k} ; \omega \!\mid\!  q_i, \mydots, q_{n+k-1}, Q_i,\mydots,Q_{i+k}) \!-\! \sum_{j=0}^{k}c(Q_{i+j})\\
	 	&\leq I(q_i ; \omega \mid  Q_i)\! +\! \mydots\! +\! I(q_{i+k} ; \omega \mid  Q_{i+k})\! -\! \sum_{j=0}^{k}c(Q_{i+j})\\
	 	&\leq (k+1)\left[I(q_i ; \omega \mid  Q^*_i) - c(Q^*_i)\right] < 0
	 	\end{align*}
	 	where the first inequality is due to the submodularity of the mutual information, and the second inequality is because $Q^*_n$ is the global maximizer of the greedy objective. The other direction of the proof is very clear: If the global optimizer is nonnegative, then querying $Q^*_n$ will not decrease the cumulative active learning reward in expectation, so stopping is not optimal.
	\end{proof}
\end{theorem}

\begin{figure*}[tbh]
	\centering
	\includegraphics[width=\textwidth]{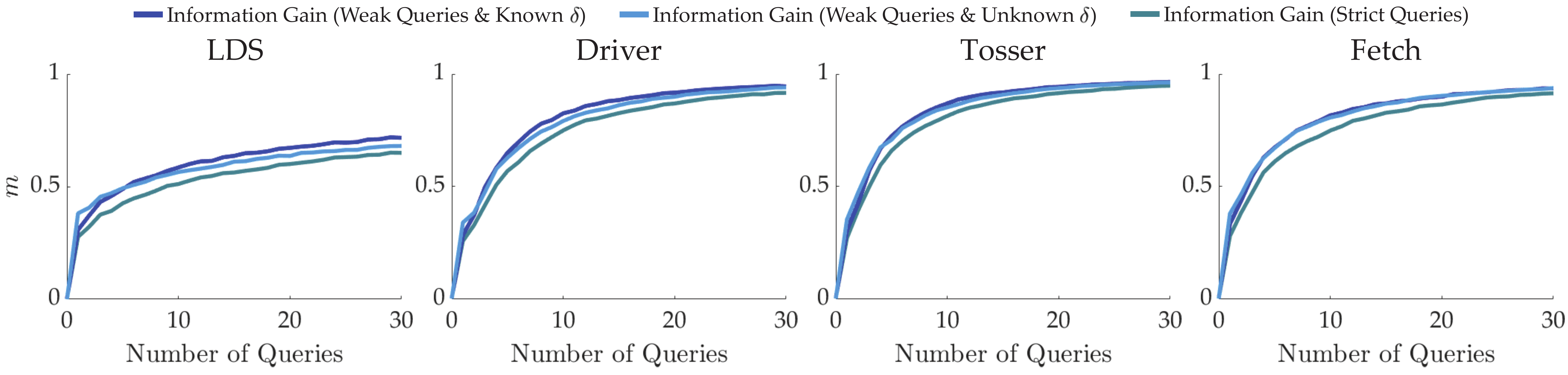}
	\caption{The simulation results with information gain formulation for unknown $\delta$. Plots are mean$\pm$s.e.}
	\label{fig:unknown_delta}
\end{figure*}

\begin{figure*}[bth]
	\centering
	\includegraphics[width=\textwidth]{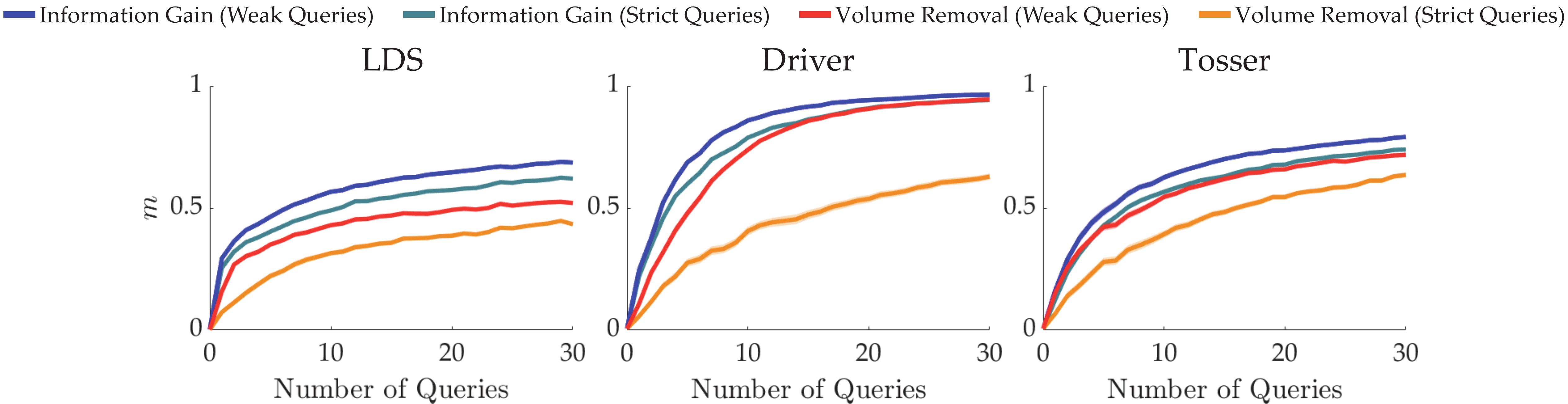}
	\caption{Alignment values are plotted (mean$\pm$s.e.) for th experiments without query space discretization, i.e., with continuous trajectory optimization for active query generation.}
	\label{fig:continuous_optimization}
\end{figure*}

\subsection{Extension to User-Specific and Unknown $\delta$}
We now derive the information gain solution when the minimum perceivable difference parameter $\delta$ of the extended human model (for weak preference queries) we introduced in the Experiments section is unknown. One can also introduce a temperature parameter $\beta$ to both standard and extended models such that $R(\xi_k)$ values will be replaced with $\beta R(\xi_k)$ in Eqs.~\eqref{eq:noisily_optimal} and \eqref{eq:weak_noisily_optimal}. This temperature parameter is useful for setting how noisy the user choices are, and learning it can ease the feature design.

Therefore, for generality, we denote all human model parameters that will be learned as a vector $\nu$. Furthermore, we denote the belief over $(\omega,\nu)$ at iteration $i$ as $b_+^i$. Since our true goal is to learn $\omega$, the optimization now becomes:
\begin{align*}
Q_i^* &= \argmax_{Q_i=\{\Lambda_1,\dots,\Lambda_K\}} \mathbb{E}_{\nu\mid Q_i,b_+^i}\left[I(q_i;\omega \mid  Q_i,b_+^i)\right]
\end{align*}
We now work on this objective as follows:
\begin{align*}
\mathbb{E}&_{\nu\mid Q_i,b_+^i}\left[I(q_i;\omega \mid  Q_i,b_+^i)\right]\\
&= \mathbb{E}_{\nu\mid Q_i,b_+^i}\Big[H(\omega \mid  \nu, Q_i,b_+^i) - \\&\qquad \mathbb{E}_{q_i\mid \nu, Q_i,b_+^i}\left[H(\omega \mid  q_i, \nu, Q_i,b_+^i)\right]\Big]\\
&= \mathbb{E}_{\nu\mid Q_i,b_+^i}\left[H(\omega \mid  \nu, Q_i,b_+^i)\right] - \\&\qquad \mathbb{E}_{\nu,q_i\mid Q_i,b_+^i}\left[H(\omega \mid  q_i, \nu, Q_i,b_+^i)\right]\\
&= -\mathbb{E}_{\nu,\omega\mid Q_i,b_+^i}\left[\log_2 P(\omega \mid  \nu, Q_i,b_+^i)\right] + \\&\qquad \mathbb{E}_{\nu,q_i,\omega\mid Q_i,b_+^i}\left[\log_2 P(\omega \mid  q_i, \nu, Q_i,b_+^i)\right]\\
&= \mathbb{E}_{\nu,q_i,\omega\mid Q_i,b_+^i}\big[\log_2 P(\omega \mid q_i, \nu, Q_i,b_+^i) - \\&\qquad \log_2 P(\omega \mid \nu, Q_i,b_+^i)\big]\\
&= \mathbb{E}_{\nu,q_i,\omega\mid Q_i,b_+^i}\big[\log_2 P(q_i \mid \omega, \nu, Q_i,b_+^i) - \\&\qquad \log_2 P(q_i \mid \nu, Q_i,b_+^i)\big]\\
&= \mathbb{E}_{\nu,q_i,\omega\mid Q_i,b_+^i} \big[\log_2 P(q_i \mid  \omega, \nu, Q_i,b_+^i) - \\&\qquad \log_2 P(\nu,q_i \mid  Q_i,b_+^i) + \log_2 P(\nu \mid  Q_i,b_+^i)\big]
\end{align*}
Noting that $P(\nu \mid  Q_i,b_+^i) = P(\nu \mid b_+^i)$, we drop the last term because it does not involve the optimization variable $Q_i$. Also noting $P(q_i \mid  \omega, \nu, Q_i,b_+^i) = P(q_i \mid  \omega, \nu, Q_i)$, the new objective is:
\begin{align*}
& \mathbb{E}_{\nu,q_i,\omega\mid Q_i,b_+^i}\!\left[\log_2 P(q_i \mid  \omega, \nu, Q_i)\!-\! \log_2 P(\nu,q_i \mid  Q_i,b_+^i)\right]\\
&\asymeq \frac1M \sum_{(\bar{\omega},\bar{\nu}) \in \Omega^+}\sum_{q_i\in Q_i}P(q_i\mid \bar{\omega},\bar{\nu},Q_i)\\&\qquad\left[\log_2 P(q_i \mid  \bar{\omega}, \bar{\nu}, Q_i)\!-\! \log_2 P(\bar{\nu},q_i \mid  Q_i,b_+^i)\right]
\end{align*}
where $\Omega^+$ is a set containing $M$ samples from $b_+^i$. Since $P(\bar{\nu},q_i \!\mid\!  Q_i,b_+^i) \!=\! \int P(q_i \!\mid\!  \bar{\nu}, \omega', Q_i)P(\bar{\nu},\omega' \!\mid\!  Q_i,b_+^i)d\omega'$ where the integration is over all possible values of $\omega$, we can write the second logarithm term as:
\begin{align*}
\log_2\left(\frac1M \sum_{\omega'\in \Omega(\bar{\nu})}P(q_i \mid  \bar{\nu},\omega', Q_i)\right)
\end{align*}
with asymptotic equality, where $\Omega(\bar{\nu})$ is the set that contains $M$ samples from $b_+^i$ with fixed $\bar{\nu}$. Note that while we can actually compute this objective, it is computationally much heavier than the case without $\nu$, because we need to take $M$ samples of $\omega$ for each $\bar{\nu}$ sample.

One property of this objective that will ease the computation is the fact that it is parallelizable. An alternative approach is to actively learn $(\omega,\nu)$ instead of just $\omega$. This will of course cause some performance loss, because we are only interested in $\omega$. However, if we learn them together, the derivation follows the derivation of Equation~\eqref{eq:IG2}, which we already presented, by simply replacing $\omega$ with $(\omega,\nu)$, and the final optimization becomes:
\begin{align*}
\argmax_{Q_i=\{\Lambda_1,\dots,\Lambda_K\}} \frac1M \sum_{q_i\in Q_i}\sum_{(\bar{\omega},\bar{\nu})\in\Omega^+} P(q_i\mid Q_i,\bar{\omega},\bar{\nu}) \\\qquad\log_2\frac{M\cdot P(q_i \mid  Q_i,\bar{\omega},\bar{\nu})}{ \sum_{(\omega',\nu')\in\Omega^+} P(q_i\mid Q_i,\omega',\nu')}
\end{align*}

\begin{figure*}[tbh]
	\centering
	\includegraphics[width=\textwidth]{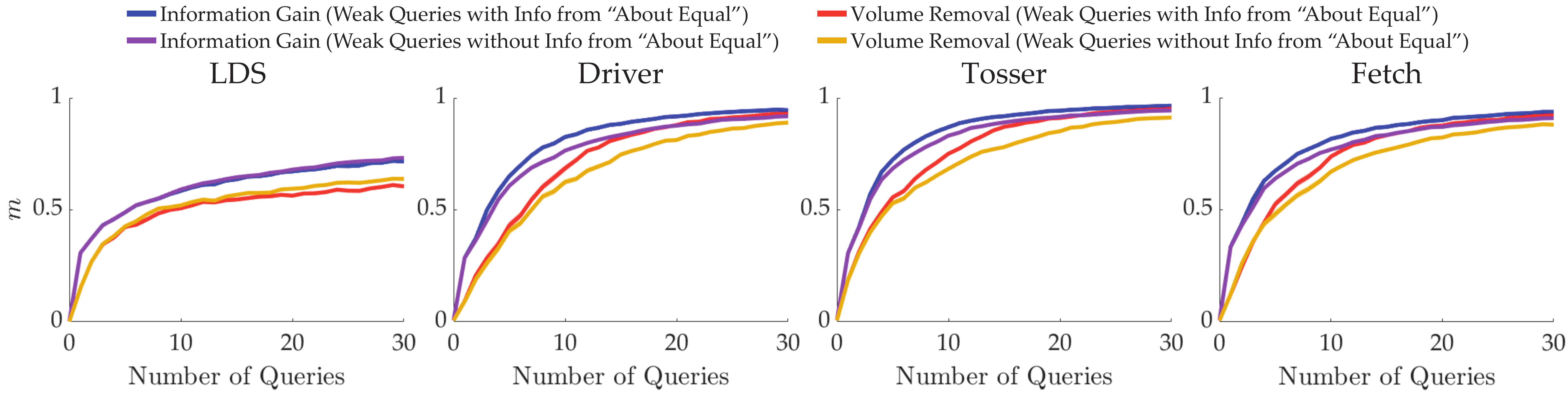}
	\caption{The results (mean$\pm$s.e.) of the simulations with weak preference queries where we use the information from ``About Equal" responses (blue and red lines) and where we don't use (purple and orange lines).}
	\label{fig:value_of_idk}
	\vspace{-5px}
\end{figure*}

\begin{figure*}[tbh]
	\centering
	\includegraphics[width=\textwidth]{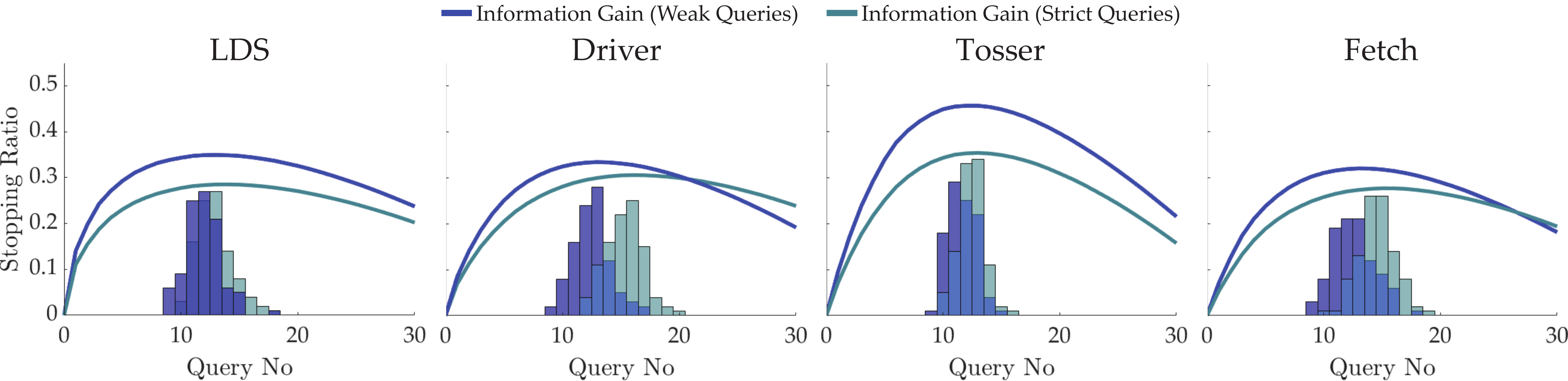}
	\caption{Simulation results for optimal stopping under query-independent costs. Line plots show cumulative active learning rewards (cumulative difference between the information gain values and the query costs), averaged over $100$ test runs and scaled for better appearance. Histograms show when optimal stopping condition is satisfied.}
	\label{fig:optimal_stopping_query_independent}
	\vspace{-5px}
\end{figure*}

Using this approximate, but computationally faster  optimization, we performed additional analysis where we compare the performances of strict preference queries, weak preference queries with known $\delta$ and weak preference queries without assuming any $\delta$ (all with the information gain formulation). As in the previous simulations, we simulated $100$ users with different random reward functions. Each user is simulated to have a true $\delta$, uniformly randomly taken from $[0,2]$. During the sampling of $\Omega^+$, we did not assume any prior knowledge about $\delta$, except the natural condition that $\delta\geq 0$. The comparison results are in Figure~\ref{fig:unknown_delta}. While knowing $\delta$ increases the performance as expected, weak preference queries are still better than strict queries even when $\delta$ is unknown. This supports the advantage of employing weak preference queries.

\subsection{Comparison of Information Gain and Volume Removal without Query Space Discretization}
We repeated the experiment that supports \textbf{H5}, and whose results are shown in Figure~\ref{fig:simulation_results_m}, without query space discretization. By optimizing over the continuous action space of the environments, we tested information gain and volume removal formulations with both strict and weak preference queries in LDS, Driver and Tosser tasks. We excluded Fetch again in order to avoid prohibitive trajectory optimization due to large action space. Figure~\ref{fig:continuous_optimization} shows the result. As it is expected, information gain formulation outperforms the volume removal with both preference query types. And, weak preference queries lead to faster learning compared to strict preference queries.

\subsection{Effect of Information from ``About Equal" Responses}
We have seen that weak preference queries consistently decrease wrong answers and improve the performance. However, this improvement is not necessarily merely due to the decrease in wrong answers. It can also be credited to the information we acquire thanks to ``About Equal" responses.

To investigate the effect of this information, we perform two additional experiments with $100$ different simulated human reward functions with weak preference queries: First, we use the information by the ``About Equal" responses; and second, we ignore such responses and remove the query from the query set to prevent repetition. Figure~\ref{fig:value_of_idk} shows the results. It can be seen that for both volume removal and information gain formulations, the information from ``About Equal" option improves the learning performance in Driver, Tosser and Fetch tasks, whereas its effect is very small in LDS.

\subsection{Optimal Stopping under Query-Independent Costs}
To investigate optimal stopping performance under query-independent costs, we defined the cost function as $c(Q) = \epsilon$, which just balances the trade-off between the number of questions and learning performance. Similar to the query-dependent costs case we described in the Experiments section, we first simulate $100$ random users and tune $\epsilon$ accordingly in the same way. We then use this tuned $\epsilon$ for our tests with $100$ different random users. Figure~\ref{fig:optimal_stopping_query_independent} shows the results. Optimal stopping rule enables terminating the process with near-optimal cumulative active learning rewards in all environments, which again supports \textbf{H9}.

\end{document}